\documentclass[10pt,twocolumn,letterpaper]{article}

\usepackage[pagenumbers]{cvpr} 

\definecolor{cvprblue}{rgb}{0.21,0.49,0.74}
\usepackage[pagebackref,breaklinks,colorlinks,allcolors=cvprblue]{hyperref}
 
\usepackage{url}
\usepackage{booktabs} 
\usepackage{multirow}
\usepackage[most]{tcolorbox}
\usepackage{contour}
\usepackage{svg}
\usepackage{bbding}
\usepackage[T1]{fontenc}
\usepackage{wrapfig}  
\usepackage{epigraph}
\usepackage[table]{xcolor}
\usepackage{caption}
\usepackage{graphicx}
\usepackage{textcomp}
\usepackage{subcaption} 
\usepackage{stfloats}
\definecolor{mylightgreen}{HTML}{D7FFD7}
\usepackage{listings}
\usepackage{cuted}
\usepackage{times}

\def\paperID{} 
\def\confName{CVPR}
\def\confYear{2026}

\title{Wow, wo, val! A Comprehensive Embodied World Model Evaluation Turing Test}

\author{
Chun-Kai Fan\textsuperscript{1,2,}\thanks{Equal contribution.}\quad
Xiaowei Chi\textsuperscript{2,3,}\footnotemark[1]\quad
Xiaozhu Ju\textsuperscript{2,}\thanks{Project leader.}\quad
Hao Li\textsuperscript{2}\quad
Yong Bao\textsuperscript{2}\\
Yu-Kai Wang\textsuperscript{1,2}\quad
Lizhang Chen\textsuperscript{1}\quad
Zhiyuan Jiang\textsuperscript{2}\quad
Kuangzhi Ge\textsuperscript{1,2}\quad
Ying Li\textsuperscript{1}\\
Weishi Mi\textsuperscript{2}\quad
Qingpo Wuwu\textsuperscript{1}\quad
Peidong Jia\textsuperscript{1,2}\quad
Yulin Luo\textsuperscript{1}\quad
Kevin Zhang\textsuperscript{1,2}\\
Zhiyuan Qin\textsuperscript{2}\quad
Yong Dai\textsuperscript{2}\quad
Sirui Han\textsuperscript{3}\quad
Yike Guo\textsuperscript{3}\quad
Shanghang Zhang\textsuperscript{1,}\thanks{Corresponding author.}\quad
Jian Tang\textsuperscript{2,}\footnotemark[3] 
\\
\small
\textsuperscript{1}State Key Laboratory of Multimedia Information Processing, School of Computer Science, Peking University\\
\small
\textsuperscript{2}Beijing Innovation Center of Humanoid Robotics\qquad
\small
\textsuperscript{3}The Hong Kong University of Science and Technology
}

\begin{document}
\maketitle
\begin{figure*}[t]
  \centering
  \includegraphics[width=\textwidth]{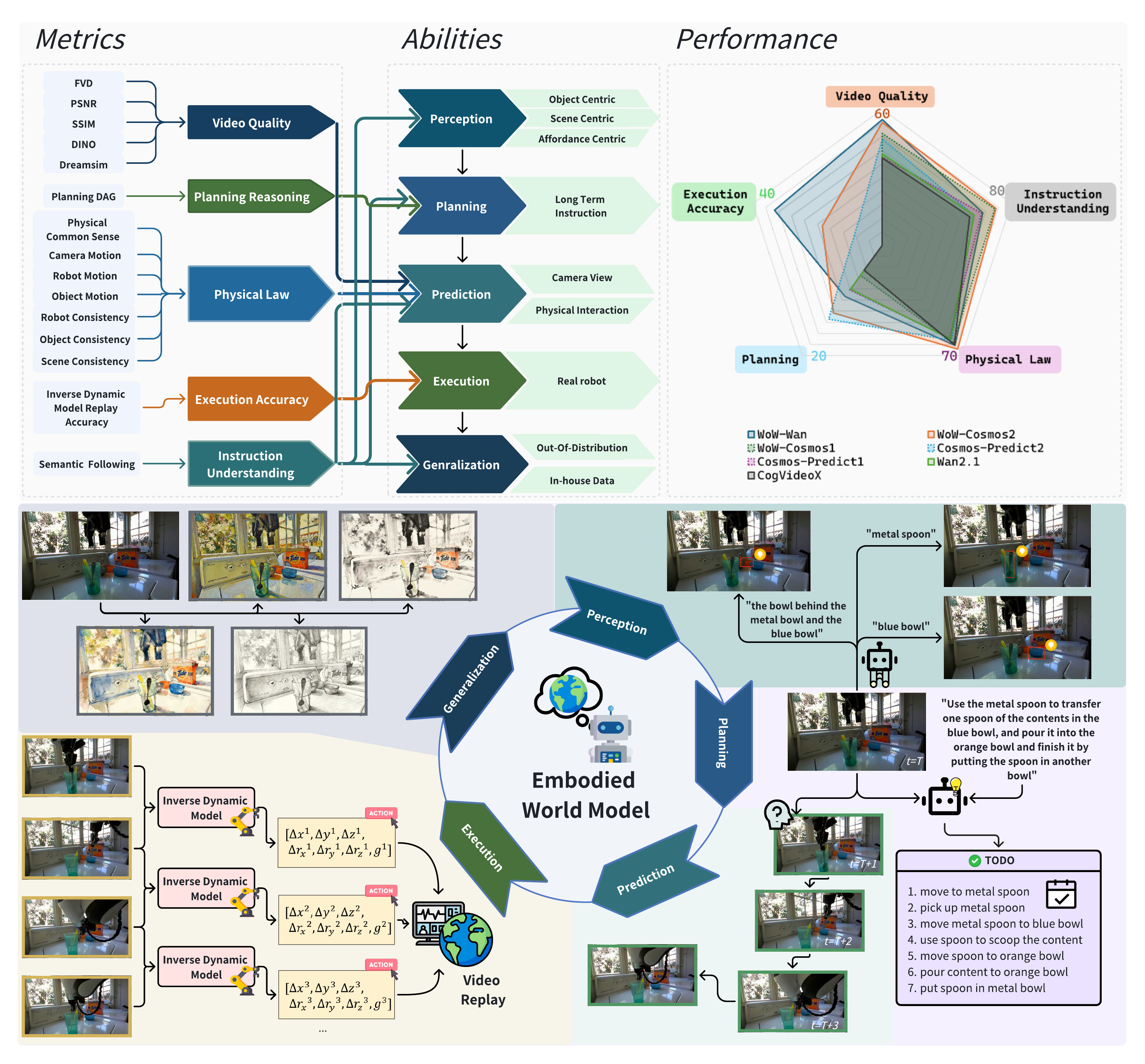} 
  \caption{\textbf{The Overview of WoW-World-Eval.} \textbf{(Top-left)} A multi-faceted \textit{Metrics} suite evaluates generated videos across five dimensions: Video Quality, Instruction Understanding, Planning Reasoning, Physical Law, and Execution Accuracy.  \textbf{(Top-center)} These metrics align with five core \textit{Abilities} of embodied world models: Perception, Planning, Prediction, Execution, and Generalization. \textbf{(Top-right)} Performance gaps across state-of-the-art models.  \textbf{(Bottom)} The benchmark follows the embodied world model pipeline from Perception to Generalization.}
  \label{fig:WoWbench_overall_design}
\end{figure*}

\begin{abstract}
As world models gain momentum in Embodied AI, an increasing number of works explore using video foundation models as predictive world models for downstream embodied tasks like 3D prediction or interactive generation.
However, before exploring these downstream tasks, video foundation models still have two critical questions unanswered: \textbf{\textit{\footnotesize (1)}} whether their generative generalization is sufficient to maintain perceptual fidelity in the eyes of human observers, and \textbf{\textit{\footnotesize (2)}} whether they are robust enough to serve as a universal prior for real-world embodied agents.
To provide a standardized framework for answering these questions, we introduce the Embodied Turing Test benchmark: \textbf{WoW-World-Eval (Wow,wo,val)}.
Building upon 609 robot manipulation data, Wow-wo-val examines five core abilities, including \textit{perception, planning, prediction, generalization, and execution}. We propose a comprehensive evaluation protocol with 22 metrics to assess the models' generation ability, which achieves a high Pearson Correlation between the overall score and human preference (>0.93) and establishes a reliable foundation for the Human Turing Test.
On Wow-wo-val, models achieve only 17.27 on long-horizon planning and at best 68.02 on physical consistency, indicating limited spatiotemporal consistency and physical reasoning. For the Inverse Dynamic Model Turing Test, we first use an IDM to evaluate the video foundation models' execution accuracy in the real world. However, most models collapse to $\approx$ 0\% success, while WoW maintains a 40.74\% success rate. These findings point to a noticeable gap between the generated videos and the real world, highlighting the urgency and necessity of benchmarking World Model in Embodied AI.
\end{abstract}
 
\section{Introduction}
\label{sec:intro}

World models – which capture an agent’s understanding of how the world changes with actions~\cite{ha2018world} – have emerged as a pivotal concept in robotics and Embodied AI. In embodied settings, a world model allows a robot to understand and predict its environment~\cite{ding2025understanding,long2025survey,fung2025embodied,li2025comprehensive}, and can function as the robot’s “internal brain”, enabling it to simulate future scenarios for planning and decision-making~\cite{chi2025wow,unifolmwma2025,wang2023anypose,Ko2023Learning}, or operate as an environment simulator~\cite{li2025vla,xiao2025world,liao2025genieenvisionerunifiedworld}.

Compared with cutting-edge spatial-prediction world models~\cite{genie3,team2025hunyuanworld}, embodied world models operate in context-rich environments that demand a deeper understanding of physical common sense. In robotics, the complexity and lack of standardization across setups further lead to a broad and diverse landscape of embodied-world-model designs. These models vary widely in their control conditions—ranging from approaches conditioned solely on images and language instructions~\cite{unifolmwma2025, chi2024eva, chi2025wow, chi2025mindlearningdualsystemworld}, to those incorporating keypoints~\cite{wang2025language}, trajectories~\cite{li2025manipdreamer3d, fu2025learning, jiang2025enerverse, quevedo2025evaluating, cen2025worldvla, guo2025ctrl}, depth, semantics, and other modalities~\cite{yang2025orv, li2025manipdreamer, zhou2024robodreamer}. They also differ in camera configurations, requiring either single-view inputs~\cite{chi2024eva, zhu2024irasim, cen2025worldvla, fu2025learning, unifolmwma2025, quevedo2025evaluating, wang2025language, chi2025mindlearningdualsystemworld, li2025manipdreamer3d} or multi-view setups~\cite{pmlr-v202-seo23a, chen2025robohorizonllmassistedmultiviewworld, pang2025learning, chi2025wow, liao2025genieenvisionerunifiedworld, yang2025orv, guo2025ctrl}. Moreover, several recent efforts have begun to pursue cross-embodiment generalization~\cite{zhi20253dflowaction, he2025scaling, chi2025wow}.
Therefore, despite the broader research of world models in general-purpose robotics, two core questions still remain: 
\begin{enumerate}
    \item Can these models generalize well enough to maintain perceptual fidelity from a human perspective?
    \item Are they robust and expressive enough to serve as universal priors for real-world embodied agents?
\end{enumerate}

Existing video-generation benchmarks~\cite{feng2024tc,ling2025vmbench,li2025worldmodelbench,yue2025ewmbench,liu2024evalcrafter,ji2024t2vbench,zheng2025vbench} largely target general-purpose settings or isolated dimensions and overlook the unique requirements of robotic world models. Most evaluations emphasize visual fidelity or coarse task success, but rarely assess deeper embodied abilities such as physical plausibility, planning rationality, and actionability. This gap makes progress difficult to measure: a model may score well on conventional video metrics~\cite{unterthiner2018towards,1284395,Oquab2023DINOv2,fardo2016formalevaluationpsnrquality} yet produce physically impossible or contextually incorrect predictions in robotic scenarios. Our results further confirm this misalignment—standard video-quality scores correlate poorly with human judgments in embodied settings—highlighting the need for more reliable evaluation standards.

Consequently, in this paper, we address these challenges by proposing a new comprehensive benchmark for embodied world models, and use it to systematically evaluate foundational models under the simplest \textit{image-to-video} setting. Our benchmark, \textbf{WoW-World-Eval}, as illustrated in Figure~\ref{fig:WoWbench_overall_design}, is designed around the core capabilities that an embodied world model should possess: perceiving the environment, understanding and planning based on task instructions, predicting and simulating future world states, executing real-world interactions, and generalizing across diverse scenarios and embodiments.
It contains about 609 robot manipulation samples with meticulous cleaning and annotation by human annotators. Also we incorporate 22 evaluation metrics aligned with our core dimensions across Video Quality, Instruction Understanding, Planning Reasoning, Physical Law, and Execution Accuracy.

Moreover, \textbf{WoW-World-Eval} follows the two-alternative forced-choice (2AFC)~\cite{fechner1860elemente} methodology from psychophysics to establish the evaluation as a standardized Turing Test for generative video models. By collecting fine-grained human answers distinguishing real and generated videos, we compute the proportion of generated videos from each model that successfully fool human evaluators. Notably, an overall human preference alignment score exceeding Pearson Correlation = 0.93 demonstrates the effectiveness of our benchmark in evaluating high-quality generations and serves as a reliable proxy for the \textit{Human Turing Test}.

In addition to the human-centered evaluation, we also introduce a \textit{Machine Turing Test}—specifically, an \textit{Inverse Dynamics Model (IDM) Turing Test}. In this setting, we assess whether videos generated by a model can “fool” an IDM that has only been trained on real-world execution sequences. If the generated videos lead the IDM to output plausible actions that are executable in the real world, it indicates that the model's outputs are indistinguishable from real data in terms of physical and action plausibility.

By evaluating existing models under this new benchmark, we reveal which models already exhibit credible world understanding and where they fall short. We believe this benchmark and the accompanying Turing Test criterion will provide a much-needed standard for the field, driving research towards embodied world models that can truly imagine the world with the accuracy and fidelity that robotics applications demand. Our contributions are threefold as follows:
\begin{itemize}
    \item A comprehensive World Model Benchmark, \textbf{WoW-World-Eval}, focuses on the Embodied AI domain, introducing a new perspective with a novel framework for the five core abilities in the embodied world model. 
    
    \item Based on \textbf{WoW-World-Eval}, we propose two novel Turing tests aligned with our benchmark, the Human Turing test and IDM Turing Test, that can distinguish the models' true ability as an embodied world model to simulate and interact with the real world.

    \item We curate 609 high-quality robot manipulation samples with careful human annotations in \textbf{WoW-World-Eval}. Using this benchmark, we conduct a comprehensive evaluation of various world models, whose performance provides new insights into the strengths, limitations, and generalization gaps of current embodied world models.
\end{itemize}

\section{Related work}
\label{sec:related}

\definecolor{GreenCheck}{rgb}{0.2, 0.6, 0.2}
\definecolor{RedCross}{rgb}{0.8, 0.2, 0.2}

\newcommand{\cmark}{\textcolor{GreenCheck}{$\checkmark$}}%
\newcommand{\xmark}{\textcolor{RedCross}{$\times$}}%

\begin{table}[!ht]
\centering
\caption{ \textbf{Comparison of benchmark features.}
A checkmark (\cmark) indicates that the benchmark explicitly includes the corresponding evaluation dimension or metric, while a cross (\xmark) indicates that it does not. 
In Core Dimensions, Perception(Percept); Prediction(Pred); Planning(Plan); Execution(Exec); Generalization(General); in Metrics Design, Video Quality(VQ); Instruction Understanding(IU); Physical Law(PL); Planning Reasoning(PR); Execution Accuracy(EA).
}

\label{tab:final_comparison}
\setlength{\tabcolsep}{2.2pt} 
\scriptsize
\begin{tabular}{@{}lcccccccccc@{}}
\toprule
& \multicolumn{5}{c}{{Core Dimensions}} &
  \multicolumn{5}{c}{{Metrics Design}} \\
\cmidrule(lr){2-6}\cmidrule(lr){7-11}
{Benchmark} &
{Percept} & {Pred} & {Plan} & {Exec} & {General} &
{VQ} & {IU} & {PL} & {PR} & {EA} \\
\midrule

Physics\textnormal{-}IQ~\cite{motamed2025generative}         & \cmark & \cmark & \xmark & \xmark & \xmark & \xmark & \xmark & \cmark & \xmark & \xmark \\
PhyGenBench~\cite{meng2024towards}                        & \cmark & \cmark & \xmark & \xmark & \xmark & \xmark & \xmark & \cmark & \xmark & \xmark \\

T2V\textnormal{-}CompBench~\cite{sun2025t2v}        & \cmark & \cmark & \xmark & \xmark & \xmark & \xmark & \cmark & \cmark & \xmark & \xmark \\
VBench\textnormal{-}2.0~\cite{zheng2025vbench}         & \cmark & \cmark & \xmark & \xmark & \cmark & \cmark & \cmark & \cmark & \xmark & \xmark \\

WorldModelBench~\cite{li2025worldmodelbench}                    & \cmark & \cmark & \xmark & \xmark & \cmark & \xmark & \cmark & \cmark & \xmark & \xmark \\
EWMBench~\cite{yue2025ewmbench}                           & \cmark & \cmark & \xmark & \xmark & \xmark & \xmark & \cmark & \cmark & \xmark & \xmark \\

{Ours}        & \cmark & \cmark & \cmark & \cmark & \cmark & \cmark & \cmark & \cmark & \cmark & \cmark \\

\bottomrule
\end{tabular}
\end{table}

The evaluation of generative world models and video generation models is rapidly evolving. Existing benchmarks can be grouped into: (1) those targeting general video quality, (2) those assessing physical reasoning, and (3) benchmarks for holistic world model evaluation. 

\textbf{Video Generation Benchmarks.} Early benchmarks mainly assessed visual fidelity and temporal coherence. TC-Bench~\cite{feng2024tc} studies temporal compositionality with transition-aware metrics. \cite{ling2025vmbench} measures perception-aligned motion quality. T2V-CompBench~\cite{sun2025t2v} evaluates compositionality via MLLM- and tracking-based methods, challenging metrics like FVD. \cite{liu2024evalcrafter,ji2024t2vbench,chi2024eva} introduce tasks related to controllability and anticipation. VBench-2.0~\cite{zheng2025vbench} focuses on “intrinsic faithfulness,” adding \textit{Physics} and \textit{Commonsense} dimensions using VLM/LLM evaluators.

\textbf{Physical Reasoning Benchmarks.} Physics-IQ~\cite{motamed2025generative} introduces physics-based diagnostics. PhyGenBench~\cite{meng2024towards} targets semantic and physical plausibility. PhysBench~\cite{chow2025physbench} evaluates VLM physical reasoning via QA tasks rather than pixel metrics. Videophy~\cite{bansal2024videophy} benchmarks physical commonsense using rule-based and learned evaluators.

\textbf{World Model Benchmarks.}
WorldModelBench~\cite{li2025worldmodelbench} and EWMBench~\cite{yue2025ewmbench} evaluate core dimensions like Perception, Prediction, and Generalization. 
Critically, as shown in Table~\ref{tab:final_comparison}, these benchmarks do not assess the crucial core dimension of Planning and Execution. Furthermore, no prior benchmark provides a metrics design that covers Planning Reasoning or Execution Accuracy in the robotics domain.
Our proposed \textbf{WoW-World-Eval} extends these efforts with a robotics-centric design, making it the most comprehensive embodied world model benchmark to date.

\section{WoW-World-Eval: A Multi-faceted Benchmark for Embodied World Models}
In this section, we introduce \textbf{WoW-World-Eval}, a novel benchmark designed to evaluate embodied world models rigorously. We formulate the core evaluation task as \textbf{conditional video generation from an initial image and a text instruction (Image-Text-to-Video)}, a setting that directly probes a model's ability to understand a given state and execute a specified action. Moving beyond metrics of visual appeal, \textbf{WoW-World-Eval} specifically assessed a model's instruction understanding, planning, observation perception abilities, and understanding of physically grounded dynamics in embodied settings. 
For introducing our benchmark, we structure our presentation by first outlining the five core capabilities that a competent embodied world model must possess (Section~\ref{ssec:core_abilities}), then detailing our data curation pipeline (Section~\ref{ssec:data_curation}), followed by our comprehensive evaluation metrics (Section~\ref{ssec:metrics}).

\begin{figure*}[t]
  \centering
  \includegraphics[width=\textwidth]{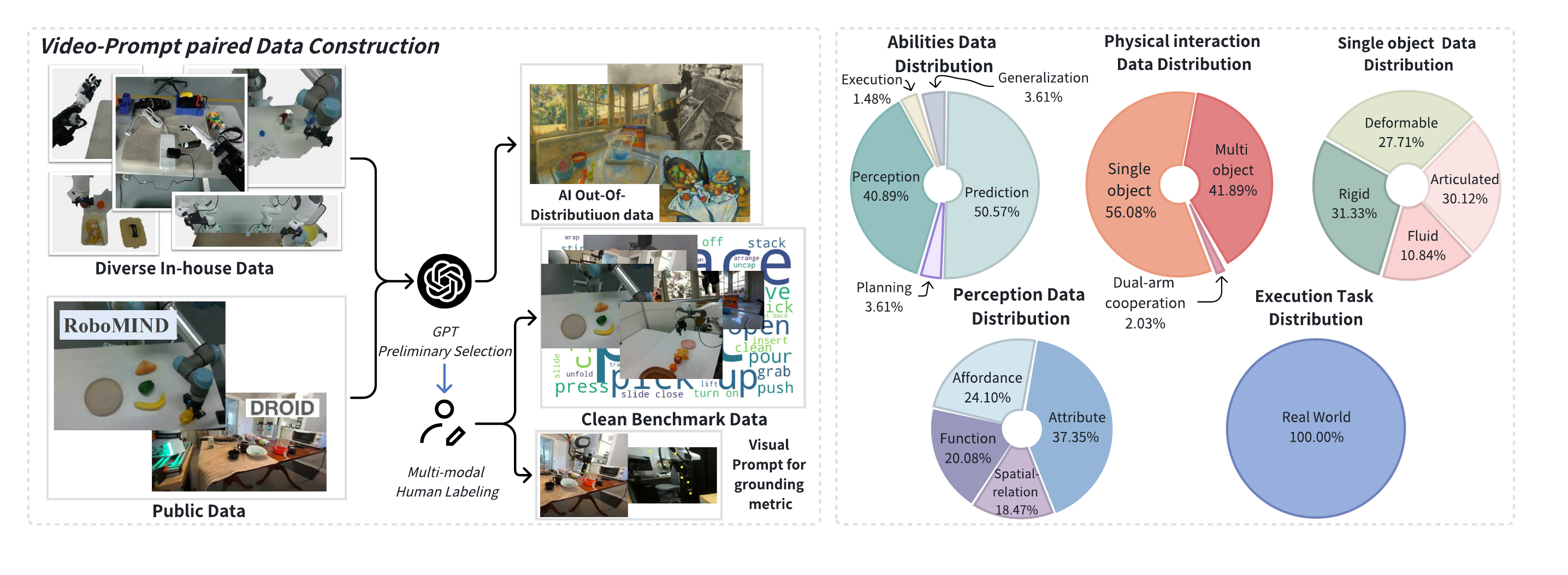} 
  \caption{\textbf{Overview of WoW-World-Eval Data Construction Pipeline and Data Statistics.} \textbf{(Left)} Public and in-house data are cleaned sequentially by GPT and human annotators to produce high-quality samples consisting of an initial image, prompt, ground-truth video, and annotated keypoints. \textbf{(Right)} Data distribution across five different dimensions from overall to fine-grained.}
  \label{fig:data_statistic}
\end{figure*}

\subsection{Core Evaluation Dimensions}
\label{ssec:core_abilities}
We posit that a truly effective embodied world model must demonstrate mastery across five fundamental and orthogonal dimensions, as outlined in our benchmark design (Figure~\ref{fig:WoWbench_overall_design}, Top-center).

\textbf{Perception Understanding.}
A world model must first accurately perceive and represent the environment to enable more reliable subsequent prediction and planning. We assess this through tasks requiring fine-grained \textbf{object recognition}~\cite{cheng2024egothinkevaluatingfirstpersonperspective, cheng2024videgothinkassessingegocentricvideo, 10654928, chow2025physbench}(attributes like color, shape, number, and size), \textbf{spatial understanding}~\cite{yang2025thinkingspacemultimodallarge, cheng2025embodiedevalevaluatemultimodalllms, song2025robospatialteachingspatialunderstanding, du2024embspatialbenchbenchmarkingspatialunderstanding} (relative positions and arrangements), and \textbf{affordance recognition}~\cite{gibson2014theory, nasiriany2025rt, cheng2025embodiedevalevaluatemultimodalllms} (identifying interactive parts of objects). 


\textbf{Decision-making and Planning.}
Embodied agents must execute long-horizon tasks~\cite{chen2024egoplanbenchbenchmarkingmultimodallarge, sermanet2023robovqamultimodallonghorizonreasoning, li2024mmromultimodalllmseligible, cheng2024videgothinkassessingegocentricvideo, yang2025embodiedbenchcomprehensivebenchmarkingmultimodal}. Therefore, we assess a model's planning ability by challenging it to generate coherent video sequences for complex instructions. This requires implicitly understanding \textbf{task decomposition}~\cite{chi2024eva, tian2025seea} into key sub-goals and respecting their \textbf{causal dependencies}. Hence, we collect 25 samples to fill in this long-term planning task, and transform the text instruction into a suitable description for the world model to plan. In order to evaluate the planning ability in the world model, we refer to the metric from RoboBench~\cite{luo2025robobench}. The detailed metric will be elaborated in section~\ref{ssec:metrics}. 


\textbf{Predictive Reasoning.}
This dimension evaluates the model's internal physics engine. Given an initial state and an action, the model must generate a future that respects core physical principles such as \textbf{object permanence}~\cite{bansal2024videophy, chow2025physbench}, \textbf{collision dynamics}~\cite{meng2024towards, chow2025physbench, bansal2024videophy, motamed2025generative}, and \textbf{trajectory plausibility}~\cite{qin2024worldsimbenchvideogenerationmodels, li2025worldmodelbench, duan2025worldscoreunifiedevaluationbenchmark, zheng2025vbench, yue2025ewmbench}. This directly probes the model's capacity to function as a world simulator~\cite{sora2024}. Therefore, we design several sub-dimensions that focus on these principles, as illustrated by the pie chart in the right of Figure~\ref{fig:data_statistic}. 


\textbf{Interactive Execution.} Interacting with the real world and executing on a real robot is the ultimate goal of an embodied world model. To examine this, we collect 9 different tasks from our in-house data that are compatible with the real robot experiments, for the model to generate. Furthermore, we will have the model generate videos based on this in-house data and use the Gripper-Centric Inverse Dynamics Model(GC-IDM)~\cite{chi2025wow}, to interpret generated videos into actions that can be executed for a real robot, finding out the \textbf{execution ability} in the world model. The detailed task names and the implementation will be described in the Appendix~\ref{ssec:IDM}.

\textbf{Generative Generalization.}
A universal world model should not only perform well on the In-Distribution data, but it should also generalize beyond the data it has seen before to demonstrate its generalization ability. For this reason, we test generalization on the in-house robot data that we collected, by using GPT-5~\cite{hurst2024gpt} to perform style transfer or image editing on it, and generating images that the world model had never seen before. We also collected some world-famous masterpiece paintings, such as \textit{"Girl with a Pearl Earring"}, and asked the world model to execute the task instructions that humans created. These two types of images constitute our \textbf{In-house Data} and \textbf{Out-of-Distribution (OOD)} dimension. 


\subsection{Data Curation Pipeline}
\label{ssec:data_curation}
To systematically evaluate these capabilities, we build a principled, semi-automated data curation pipeline (Figure~\ref{fig:data_statistic} Left). Our dataset combines open-source robotics data (e.g., RoboMIND~\cite{wu2025robomind}, DROID~\cite{khazatsky2024droid}), in-house trajectories, and AI-generated OOD samples to ensure coverage and diversity. GPT-4o~\cite{hurst2024gpt} is initially used as an intelligent annotator, scoring the matching level of video–instruction pairs based on our five capability dimensions and their subdivisions, specifically examining which instructions match which dimensions, thereby achieving large-scale filtering and coarse categorization. Human experts then verify all samples to guarantee category accuracy and resolve edge cases. Five additional annotators selected the best initial frames for generation (both the robotic arm and the manipulated object were in the same frame) and key point annotations (the robotic arm gripper, joints, and the manipulated object) on the initial frame for evaluation metrics.

Each benchmark entry contains: (1) a natural-language instruction, (2) an initial image, (3) a ground-truth video, and (4) annotated keypoints (in Prediction). In total, the dataset comprises 609 samples across all dimensions (Figure~\ref{fig:data_statistic} Right). Prediction (50.57\%) and Perception (40.89\%) dominate the ability distribution. The perception subdivisions includes object attribute, object affordance, object function, and spatial-relation tasks (249 samples). Physical interaction covers single-object manipulation (56.08\%), multi-object interaction (41.89\%), and dual-arm cooperation (2.03\%). We further include 107 non-occluded views and 54 semi-occluded views in the initial frames to test robustness. With execution data, we choose 9 real-world tasks from easy to hard for the settings.

\subsection{Multi-faceted Evaluation Metrics}
\label{ssec:metrics}
Our evaluation protocol is a suite of metrics designed to be as comprehensive as the capabilities we measure. We introduce several novel metrics alongside standard ones, grouped by the property they assess. For detailed information of all the metrics, see Appendix~\ref{sec:more_metrics}.

\textbf{Visual Fidelity.}
To fully evaluate visual fidelity, we report standard video quality metrics spanning pixel-, perceptual-, and distribution-level quality perspectively: \textbf{PSNR}\cite{fardo2016formalevaluationpsnrquality} measures pixel-level fidelity via the log ratio between the maximum signal value and the mean-squared error. To capture higher-level content consistency beyond pixel statistics, we compute \textbf{SSIM}\cite{1284395} for a structural measure and \textbf{DINO}~\cite{Oquab2023DINOv2} as a semantic/instance-alignment signal, combine with \textbf{Dreamsim}~\cite{fu2023dreamsim}, a human-aligned perceptual similarity metric trained from human triplet judgments. \textbf{FVD}\cite{unterthiner2018towards} compares the distributions of real and generated videos by computing a Fréchet distance, which reflecting overall realism and temporal dynamics at the dataset level. 


\textbf{Instruction Semantic Alignment.}
We use GPT-4o~\cite{hurst2024gpt} as a scalable evaluator to assess semantic alignment between the given instruction and the generated videos. Depending on whether ground-truth (GT) video is available, we adopt two methods:

\begin{itemize}
    \item \textbf{With Ground-Truth:} We first prompt GPT-4o to extract structured descriptions (Initial-, Processing-, Final-state) from both the generated and GT videos. A vision–language model then scores their \textbf{Caption Score}, which scales from 1-to-5. In addition, GPT-4o also evaluates the generated video action-object pairs against the instruction to produce a \textbf{Sequence Match Score} (0–1 scale), which stands for the correctness of the order of the action-object pairs, and an \textbf{Execution Quality Score} (1–5 scale) for the correctness of the action-object pairs. \emph{We report all three metrics in this setting.}
    \item \textbf{Without Ground-Truth (in Generalization):} When GT video is unavailable, we only assess instruction adherence: GPT-4o directly analyzes the generated video and the instruction to output the \textbf{Sequence Match Score} and the \textbf{Execution Quality Score}. \emph{For this setting, only these two metrics are reported.}
\end{itemize}

\textbf{Physical Consistency and Causal Reasoning.}
To quantify the physical plausibility of generated videos, we compute three kinds of metrics as follows:
\begin{itemize}
    \item \textbf{Mask-guided Regional Consistency.}
    To better disentangle inconsistencies in the background, robot arm, and manipulated object, we propose \textbf{Mask-guided Regional Consistency}. We first use the GroundedSAM2~\cite{ren2024grounded} with human annotation to obtain masks for the robot arm, the manipulated object(s), and the background in each frame. We then compute region-specific embeddings using a vision foundation model (e.g., DINOv3~\cite{siméoni2025dinov3}) and measure cosine similarity across time for each region separately. This allows us to pinpoint the source of temporal flaws—for instance, identifying a "jittery" robot arm even when the object and background are stable.

    \item \textbf{Trajectory Consistency:} For comparing the trajectory between generated videos and ground-truth counterparts, we track both the end-effector and object trajectories. In our benchmark, we leverage SAM2  ~\cite{ravi2024sam}, given a few representative key points in the initial frame that humans annotated, to follow the motion of objects in both videos. 
    Trajectory similarity is then evaluated using a complementary set of metrics: \textbf{Mean Euclidean Distance (MED)}~\cite{Dokmanic_2015} to capture average deviation, \textbf{Dynamic Time Warping (DTW)}~\cite{Müller2007} to assess temporal alignment, and \textbf{Fréchet Distance}~\cite{Eiter1994ComputingDF} to measure worst-case path similarity.
    
    \item \textbf{Physical common sense:} Physical common sense covers dimensions ranging from object interaction and properties to temporal consistency, lighting, fluid dynamics, and local anomalies. To automatically score these six distinct dimensions, we fine-tuned Qwen-2.5-VL~\cite{bai2025qwen25vltechnicalreport} using a two-stage GRPO~\cite{shao2024deepseekmath} process. First, we used several physical and temporal datasets~\cite{krojer2025shortcutawarevideoqabenchmarkphysical, chow2025physbench, bansal2025videophy, cores2025losttimenewtemporal, liu2024tempcompassvideollmsreally, zhang2024unveiling} for an initial GRPO fine-tuning to enhance the model's understanding of video content, temporal causality, and physical laws. Second, we used a human-annotated dataset of 1,297 ratings (a mix of synthetic and real videos) for further training, aligning the model to human preferences. The fine-tuned model was finally used to score these dimensions on a 1-to-5 scale.

\end{itemize}

\textbf{Planning and Task Decomposition.}
\label{ssec:dag}
To evaluate long-horizon planning, we refer to the metric of RoboBench~\cite{luo2025robobench} based on Directed Acyclic Graphs (DAGs). We first parse the natural language instruction and ground-truth video into a ground-truth plan DAG, where nodes are atomic actions parameterized by \(\langle \text{skill}, \text{object}, \text{args} \rangle\) and edges represent dependencies. This representation flexibly handles non-unique but valid action orderings. We then compare the model-generated plan (which also uses the same approach from the video) to the ground-truth DAG using two scores:
\begin{enumerate}
    \item \textbf{Node Correctness:} The fraction of correctly predicted nodes aligns with the ground-truth nodes.
    \item \textbf{Task Completion:} Using the MLLM to conduct a lightweight world-simulation rollout using: (i) first-frame image, (ii) reference action list, and (iii) reference DAG, and record the stage changes that model-predicted DAGs have correctly executed.
\end{enumerate}
The final planning score \textbf{LongHorizon} integrates these aspects to reward both correctness and completeness:
\[
S_{\text{LongHorizon}} = (S_{\text{NodeCorrectness}} + S_{\text{TaskCompletion}})*50.
\]

\subsection{Overall Benchmark Score}
\textbf{Setup.}
For each model $i$ and metric $m$, we map the raw measurement $x_{i,m}$ to a common desirability score
$s_{i,m}\in(0,100)$ via a monotone parametric mapping applied after an absolute pre-scaling to $[0,1]$.
We then aggregate desirability scores by weighted arithmetic means at both metric-group and overall levels.

\textbf{Pre-scale to $[0,1]$ with absolute anchors.}
Let $L_m<U_m$ be fixed anchors for metric $m$ (documented per-metric).
Define the clipping operator $\mathrm{clip}(u;a,b)=\min\{\max\{u,a\},\,b\}$.
We first clamp raw values to $[L_m,U_m]$, then linearly map to $[0,1]$; for “higher-is-better” (HIB) metrics:
\[
\hat x_{i,m}^{\mathrm{HIB}}
=\frac{\mathrm{clip}(x_{i,m};L_m,U_m)-L_m}{U_m-L_m}\in[0,1],
\]
And for “lower-is-better” (LIB) metrics:
\[
\hat x_{i,m}^{\mathrm{LIB}}
=1-\hat x_{i,m}^{\mathrm{HIB}}\in[0,1].
\]
We use absolute anchors for two common metrics:
\[
\textbf{PSNR (HIB):}\quad L_{\text{PSNR}}=0,\ U_{\text{PSNR}}=50\
\]
\[
\textbf{FVD (LIB):}\quad L_{\text{FVD}}=0,\ U_{\text{FVD}}=2000\ .
\]
For other metrics, $(L_m,U_m)$ are fixed per protocol (e.g., theoretical bounds for bounded scales, task-specific absolute targets for unbounded ones).

\textbf{Monotone parametric mappings.}
After pre-scaling, we apply a single-parameter monotone transform $f_m(\cdot;\theta_m)$ and then rescale to $(0,100)$:
\[
s_{i,m}=100\,f_m\!\big(\hat x_{i,m};\theta_m\big),\qquad s_{i,m}\in(0,100).
\]
We consider the following families (all are strictly increasing on $[0,1]$):
\[
\begin{aligned}
\textbf{Simple:} &\quad f_(x) = x,\\[3pt]
\textbf{Power (Gamma):} &\quad f_\gamma(x) = x^\gamma,\ \gamma>0,\\[3pt]
\textbf{Logit temperature:} &\quad
f_T(x) = \sigma(\mathrm{logit}(x)/T),\ T>0,\\
&\quad \sigma(t)=\tfrac{1}{1+e^{-t}},\\[3pt]
\textbf{Tanh slope:} &\quad
f_\kappa(x)=\tfrac12(\tanh(\kappa(2x-1))+1),\\ 
&\quad \kappa>0.
\end{aligned}
\]
Particularly detailed information can be found in the Appendix~\ref{ssec:mappings}
.



\textbf{Aggregation (weighted arithmetic mean).}
Metrics are grouped into categories $g$ (e.g., \emph{quality}, \emph{instruction}, \emph{physical}, \emph{planning}).
For each model $i$, let $\mathcal{M}_g$ denote the set of metrics available in group $g$, and
$|\mathcal{M}_g|=N_{i,g}>0$ be its cardinality.
We aggregate first within each group by averaging and then across groups with normalized weights.

Let nonnegative group weights $\{W_g\}$ sum to one over the groups available for model $i$.
The overall score is
\[
O_i=\sum_{g\in\mathcal{G}_i}
\frac{W_g}{\sum_{h\in\mathcal{G}_i} W_h}
\left(
    \frac{1}{|\mathcal{M}_g|}
    \sum_{m\in\mathcal{M}_g} s_{i,m}
\right).
\]
where $\mathcal{G}_i=\{g:\,N_{i,g}>0\}$ is the set of groups for which model $i$ has at least one valid metric.
Setting $W_g\equiv 1$ yields an unweighted overall mean across all available groups.

\section{Experiments}
\label{sec:experiments}
\textbf{Models.} We evaluate our benchmark across a diverse set of commercial closed-source, open-source video generation models, and the World Foundation Model using direct Image-Text-to-Video generation. The closed-source models include Kling-2.1~\cite{Kling} and Hailuo I2V-02~\cite{Hailuo}, two state-of-the-art proprietary systems optimized for zero-shot text-to-video and image-to-video generation. For open-source baselines, we include CogVideoX1.5-I2V-5B~\cite{yang2024cogvideox}, one of the earliest large-scale video diffusion models, and Wan2.1-I2V-14B~\cite{wan2025wan}, an advanced text-to-video generator emphasizing temporal coherence. Furthermore, we evaluate the World Foundation Model, Cosmos-Predict1-7B~\cite{agarwal2025cosmos} and Cosmos-Predict2-2B~\cite{cosmos2}, which focus on physics, scene, and world modeling under multimodal prompts. Finally, we evaluate the Embodied World Foundation Model called WoW~\cite{chi2025wow}, which was trained for physically consistent and instruction-aware video generation in the embodied domain on different backbones. Together, these models cover a broad spectrum of architectures and training paradigms, enabling a comprehensive comparison across diverse dimensions in our benchmark.

\begin{table*}[t]
\centering
\caption{\textbf{Comparative analysis of different video generation models in Video Quality, Instruction Understanding and Planning Reasoning.} 
All metrics: higher is better. Best results are \textbf{bold} with green highlight.
Seq. stands for Sequence; Exec. stands for Execution.}
\label{tab:foundation-model-comparison-VQ-IU-P}
\resizebox{\textwidth}{!}{
\begin{tabular}{lccccccccccc}
\toprule
\multirow{2}{*}{\textbf{Model}}
& \multicolumn{6}{c}{\textbf{Video Quality}} 
& \multicolumn{4}{c}{\textbf{Instruction Understanding}} 
& \textbf{Planning Reasoning}\\
\cmidrule(lr){2-7} 
\cmidrule(lr){8-11}
\cmidrule(lr){12-12}
& \textbf{FVD} & \textbf{PSNR} & \textbf{SSIM} & \textbf{DINO} & \textbf{DreamSim} & \textbf{Overall} 
  & \textbf{Caption Score} & \textbf{Seq. Match Score} & \textbf{Exec. Quality Score} & \textbf{Overall} 
  & \textbf{Planning DAG}\\
\midrule
\rowcolor{lightgray}
\multicolumn{12}{c}{\textbf{Closed-Source Model}} \\
\midrule
\textbf{Kling} & 80.41 & 5.41 & 85.00 & 36.95 & 58.43 & 53.24 & 81.91 & 0.25 & 1.75 & 27.97 & 2.50 \\
\textbf{Hailuo} & 85.01 & \cellcolor{mylightgreen}\textbf{6.06} & \cellcolor{mylightgreen}\textbf{85.97} & \cellcolor{mylightgreen}\textbf{41.45} & 61.95 & \cellcolor{mylightgreen}\textbf{56.09} & \cellcolor{mylightgreen}\textbf{88.78} & 51.07 & \cellcolor{mylightgreen}\textbf{70.48} & 70.11 & \cellcolor{mylightgreen}\textbf{17.27} \\
\midrule
\rowcolor{lightgray}
\multicolumn{12}{c}{\textbf{Open-Source Model}} \\
\midrule
\textbf{CogVideoX} & 50.49 & 2.48 & 80.48 & 17.06 & 42.09 & 38.52 & 79.66 & 38.44 & 44.18 & 54.09 & 4.55 \\
\textbf{Cosmos-Predict1} & 62.99 & 2.91 & 79.86 & 15.62 & 33.92 & 39.06 & 83.60 & 50.89 & 49.89 & 61.46 & 8.10 \\
\textbf{Wan2.1} & 62.95 & 1.85 & 74.09 & 20.50 & 41.76 & 40.23 & 82.28 & 41.43 & 46.84 & 56.85 & 7.95 \\
\textbf{Cosmos-Predict2} & 78.77 & 2.27 & 75.69 & 24.20 & 53.12 & 46.81 & 84.16 & 55.51 & 30.72 & 56.80 & 13.41 \\
\textbf{WoW-cosmos1} & 83.03 & 4.88 & 83.22 & 21.89 & 53.71 & 49.35 & 85.34 & \cellcolor{mylightgreen}\textbf{63.33} & 60.38 & 69.68 & 5.45 \\
\textbf{WoW-wan} & 84.89 & 5.95 & 85.76 & 37.74 & 62.57 & 55.38 & 84.85 & 52.45 & 49.18 & 62.16 & 9.32 \\
\textbf{WoW-cosmos2} & \cellcolor{mylightgreen}\textbf{85.71} & 3.52 & 78.33 & 38.63 & \cellcolor{mylightgreen}\textbf{64.42} & 54.12 & 85.85 & 59.04 & 66.18 & \cellcolor{mylightgreen}\textbf{70.36} & 12.27 \\
\bottomrule
\end{tabular}
}
\end{table*}

\begin{table*}[t]
\centering
\caption{\textbf{Comparative analysis of different video generation models in Physical Law.}
Con. stands for Consistency; Traj. stands for Trajectory; Obj. stands for Object; Cam. stands for Camera.}
\label{tab:foundation-model-comparison-PL}
\resizebox{\textwidth}{!}{
\begin{tabular}{lcccccccccccccc}
\toprule
\multirow{2}{*}{\textbf{Model}} 
& \multicolumn{13}{c}{\textbf{Physical Law}} 
& \multirow{2}{*}{\textbf{Overall}}  \\
\cmidrule(lr){2-14}
& \textbf{Robot Con.} & \textbf{Obj. Con.} & \textbf{Scene Con.} & \textbf{Robot Traj. L2norm} & \textbf{Robot Traj. DTW} & \textbf{Robot Traj. FD} & \textbf{Obj. Traj. L2norm} & \textbf{Obj. Traj. DTW} & \textbf{Obj. Traj. FD} & \textbf{Physical Score} & \textbf{Cam. ATE} & \textbf{Cam. RPE} & \textbf{Overall} \\
\midrule
\rowcolor{lightgray}
\multicolumn{15}{c}{\textbf{Closed-Source Model}} \\
\midrule
\textbf{Kling} & \cellcolor{mylightgreen}\textbf{57.62} & 16.09 & 89.26 & 56.08 & 86.05 & 25.27 & 87.14 & 87.31 & 41.77 & 71.72 & 98.12 & 99.86 & \cellcolor{mylightgreen}\textbf{68.02} & 37.93 \\
\textbf{Hailuo} & 56.99 & 10.08 & \cellcolor{mylightgreen}\textbf{89.95} & 56.79 & 86.53 & 22.79 & \cellcolor{mylightgreen}\textbf{87.87} & \cellcolor{mylightgreen}\textbf{88.96} & \cellcolor{mylightgreen}\textbf{44.90} & 57.64 & 98.26 & 99.86 & 66.72 & \cellcolor{mylightgreen}\textbf{52.55} \\
\midrule
\rowcolor{lightgray}
\multicolumn{15}{c}{\textbf{Open-Source Model}} \\
\midrule
\textbf{CogVideoX} & 59.17 & 5.82 & 82.12 & 47.67 & 81.31 & 16.62 & 83.69 & 83.14 & 32.16 & 70.51 & 97.55 & 99.85 & 63.30 & 40.12 \\
\textbf{Cosmos-Predict1} & 39.02 & 4.40 & 71.69 & 54.87 & 84.60 & 23.55 & 83.93 & 83.48 & 36.99 & 30.21 & 96.23 & 99.66 & 59.05 & 41.93 \\
\textbf{Wan2.1} & 38.57 & 4.72 & 78.38 & 46.61 & 80.51 & 14.57 & 81.91 & 81.98 & 30.71 & \cellcolor{mylightgreen}\textbf{74.98} & 83.69 & 99.36 & 59.66 & 41.18 \\
\textbf{Cosmos-Predict2} & 34.61 & 5.83 & 68.34 & 53.15 & 84.99 & 22.24 & 84.82 & 84.92 & 40.74 & 48.78 & 98.41 & \cellcolor{mylightgreen}\textbf{99.87} & 60.56 & 44.40 \\
\textbf{WoW-cosmos1} & 44.72 & 13.46 & 77.78 & 56.00 & 86.77 & 25.10 & 86.00 & 86.49 & 37.86 & 37.10 & 96.42 & 99.66 & 62.28 & 46.70 \\
\textbf{WoW-wan} & 50.27 & 9.83 & 87.10 & 53.61 & 85.15 & 21.19 & 85.84 & 86.46 & 40.59 & 47.96 & 97.21 & 99.73 & 63.75 & 47.66 \\
\textbf{WoW-cosmos2} & 49.01 & \cellcolor{mylightgreen}\textbf{16.76} & 85.20 & \cellcolor{mylightgreen}\textbf{59.97} & \cellcolor{mylightgreen}\textbf{87.56} & \cellcolor{mylightgreen}\textbf{30.53} & 86.93 & 87.06 & 44.46 & 48.17 & \cellcolor{mylightgreen}\textbf{98.64} & \cellcolor{mylightgreen}\textbf{99.87} & 66.18 & 50.74 \\
\bottomrule
\end{tabular}
}
\end{table*}

\subsection{Quantitative Evaluation Results}
\label{ssec:quantitative}
As shown in Figure~\ref{fig:WoWbench_overall_design} (Top-left and center), each ability dimension is evaluated by a targeted subset of metrics. We benchmark core dimensions with four categories of metrics—Video Quality, Instruction Understanding, Physical Law, and Planning Reasoning—with all scores normalized to a 0–100 scale. Results are summarized in Tables~\ref{tab:foundation-model-comparison-VQ-IU-P} and~\ref{tab:foundation-model-comparison-PL}.

\textbf{Overall Performance.} Overall score balanced the four capabilities. Under this score, the best closed-source model, Hailuo, achieves the highest score (52.55). Among open-source models, WoW-cosmos2 consistently ranks highest, reaching 50.74, with balanced improvements in Video Quality, Instruction Understanding, and Physics Law, whereas earlier baselines, such as CogVideoX, lag in stability and physical consistency. The importance of overall balance is further highlighted by Kling, which is strong in Video Quality and achieves the best Physical Law score, yet attains a low score (37.93) due to severe deficiencies in Instruction Understanding and Planning. In conclusion, these four dimensions are indispensable for embodied world model benchmarks.

\textbf{Video Quality.} In Table~\ref{tab:foundation-model-comparison-VQ-IU-P}, on the Video Quality axis, Hailuo obtains the highest closed-source overall score (56.09), while WoW-wan leads the open-source group, achieving a nearly matched performance (55.38). WoW-cosmos2 attains the best DreamSim (64.42) and the strongest FVD (85.71), even slightly exceeding Hailuo. These results suggest that WoW mainly focuses on high-level perceptual and distributional realism. For commercial models, they gain an excellent balance between low-level fidelity and high-level perceptual and distributional realism.

\textbf{Instruction Understanding.} For Instruction Understanding, WoW-cosmos2 achieves the best overall score among open-source models (70.36), surpassing the strongest closed-source model, Hailuo (70.11). When decomposing the metrics, we observe complementary strengths across sub-dimensions. Hailuo excels at Caption Score (88.78) and Exec. Quality (70.48), while WoW-cosmos1 shows the best Seq. Match (63.33), highlighting that both of them benefit from large-scale text–video alignment for instruction comprehension. Especially, WoW-cosmos2 better enforces procedural consistency—i.e., correctly ordering and realizing multi-step instructions over time—while remaining competitive in execution quality.

\textbf{Physical Law.} Table~\ref{tab:foundation-model-comparison-PL} shows that closed-source models establish the upper bound on overall score, with Kling and Hailuo scoring 68.02 and 66.72, respectively. Among open-source models, WoW-cosmos2 leads open-source models (66.18), with strong robot trajectory and camera-motion consistency. Firstly, from consistency perspective, the differentiation in Scene Consistency is not as intense as other two, which means all models performed well across this consistency. Secondly, WoW-cosmos2 attains the strongest Robot Trajectory and Hailuo attains the strongest Object Trajectory, indicates that commercial models focus more on object manipulation, while WoW-cosmos2 is more accurate in handling robot arm movements. Thirdly, camera motion is basically saturated; that is, the camera is very stable across all models, and sudden, erratic movements are rare. The last, these results all show that open-source models closely track commercial models in physical realism.

\textbf{Planning Reasoning.} Planning Reasoning remains a primary bottleneck for current world models. As shown in Table~\ref{tab:foundation-model-comparison-VQ-IU-P}, Planning DAG scores occupy a markedly lower and more compressed range than others. The best performance is achieved by Hailuo (17.27), whereas Cosmos-Predict2 unexpectedly leads the open-source group (13.41), followed by WoW-cosmos2 (12.27) and WoW-wan (9.32). Models such as CogVideoX and Kling exhibit weak long-term planning, indicating that long-horizon planning and temporal reasoning still represent underexplored and require more explicit planning representations or control.

\subsection{Dense prompts quantitative results}
\begin{table}[!h]
\centering
\caption{\textbf{Comparative analysis of different video generation models with dense prompts.}
All metrics: higher is better. Best results are \textbf{bold} with green highlight.}
\label{tab:dense-comparison}
\setlength{\tabcolsep}{3pt} 
\begin{tabular}{lccccc}
\toprule
\textbf{Model} & \textbf{VQ} & \textbf{IU} & \textbf{PL} & \textbf{PR} & \textbf{Overall} \\
\midrule
\textbf{Cosmos-Predict1} & 35.48 & 61.07 & 53.78 & 7.5 & 39.46 \\
\textbf{Cosmos-Predict2} & 49.75 & 75.96 & 64.66 & \cellcolor{mylightgreen}\textbf{13.86} & 51.06 \\
\textbf{WoW-cosmos1} & 59.41 & 72.54 & \cellcolor{mylightgreen}\textbf{69.71} & 11.14 & \cellcolor{mylightgreen}\textbf{53.20} \\
\textbf{WoW-wan} & \cellcolor{mylightgreen}\textbf{60.55} & 50.83 & 67.48 & 8.64 & 46.88 \\
\textbf{WoW-cosmos2} & 56.86 & \cellcolor{mylightgreen}\textbf{76.16} & 67.15 & 9.09 & 52.32 \\
\bottomrule
\end{tabular}
\end{table}

Prior work~\cite{fan2025instancecap} has shown that the textual prompt has a substantial impact on video generation quality, particularly for models that rely on language-conditioned. In our setting, Image-Text-to-Video generation—the initial frame is fixed and cannot be altered, making the text prompt the primary controllable input. Our main results, Table~\ref{tab:foundation-model-comparison-VQ-IU-P} and~\ref{tab:foundation-model-comparison-PL}, therefore, use concise prompts that specify only the task goal, without detailed descriptions of the scene, or step-by-step execution. Such under-specified prompts may limit the model’s ability to ground the intended task semantics.

To study this effect, we use InternVL3-78B~\cite{zhu2025internvl3} to expand each short prompt into a dense prompt containing explicit environment and object references, sub-goals, and physical constraints. We then re-evaluate models on \textbf{WoW-World-Eval} using dense prompts in the above Table~\ref{tab:dense-comparison}.

Across all evaluation dimensions, dense prompts lead to consistent and sometimes substantial improvements. In \textbf{Video Quality}, every model benefits from richer textual conditioning, with the best embodied world model surpassing the closed-source commercial baseline (60.55 vs. 56.09). \textbf{Instruction Understanding} exhibits even larger gains: nearly all models improve, and Cosmos-Predict2 increases by almost 20, although WoW-cosmos2 remains the most stable performer across both prompt types. Improvements in \textbf{Physical Law} are similarly widespread, with WoW-cosmos1 showing the largest jump (+7), surpassing the previous short-prompt SOTA. In contrast, in \textbf{Planning Reasoning}, DAG metric shows minimal change, indicating that detailed prompting alone cannot compensate for the current limitations of world models in long-horizon reasoning and structured task decomposition. Overall, dense prompts provide richer semantic and physical cues that meaningfully enhance generation quality, while exposing fundamental gaps in planning capability that require deeper architectural advances beyond prompt refinement.

\begin{figure*}[!t]
  \centering
  \begin{subfigure}{0.5\textwidth}
    \centering
    \includegraphics[width=\linewidth]{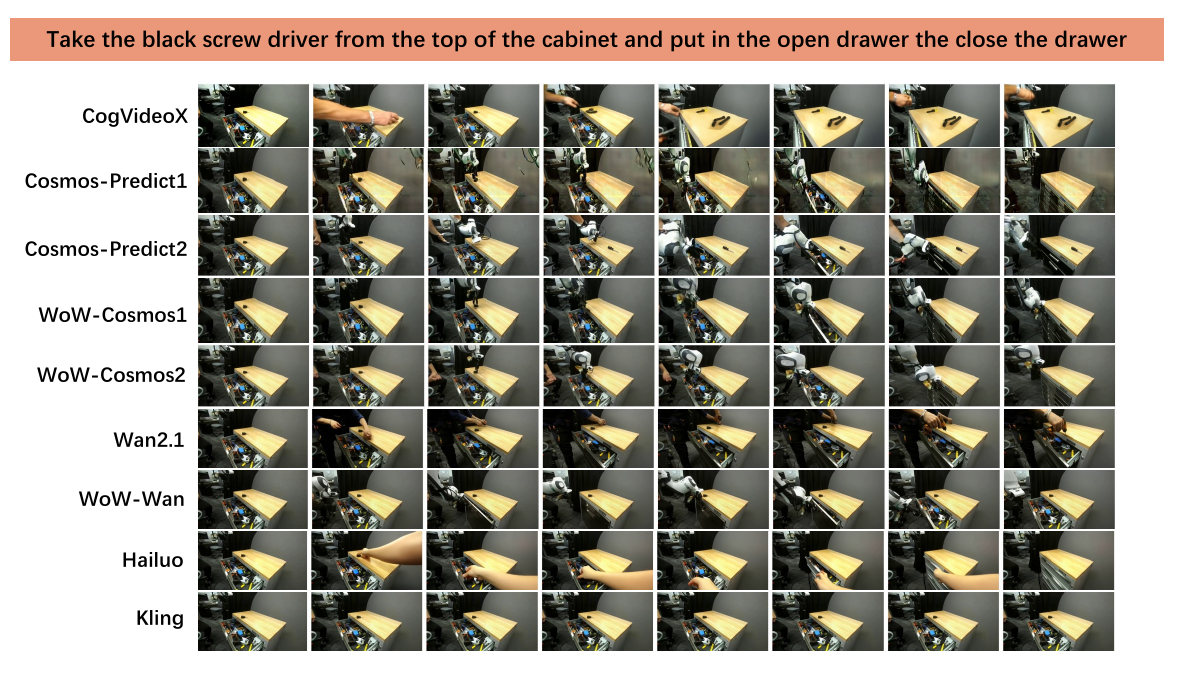}
    \caption{Qualitative Evaluation Example.}
    \label{fig:visualization1}
  \end{subfigure}\hfill
  \begin{subfigure}{0.5\textwidth}
    \centering
    \includegraphics[width=\linewidth]{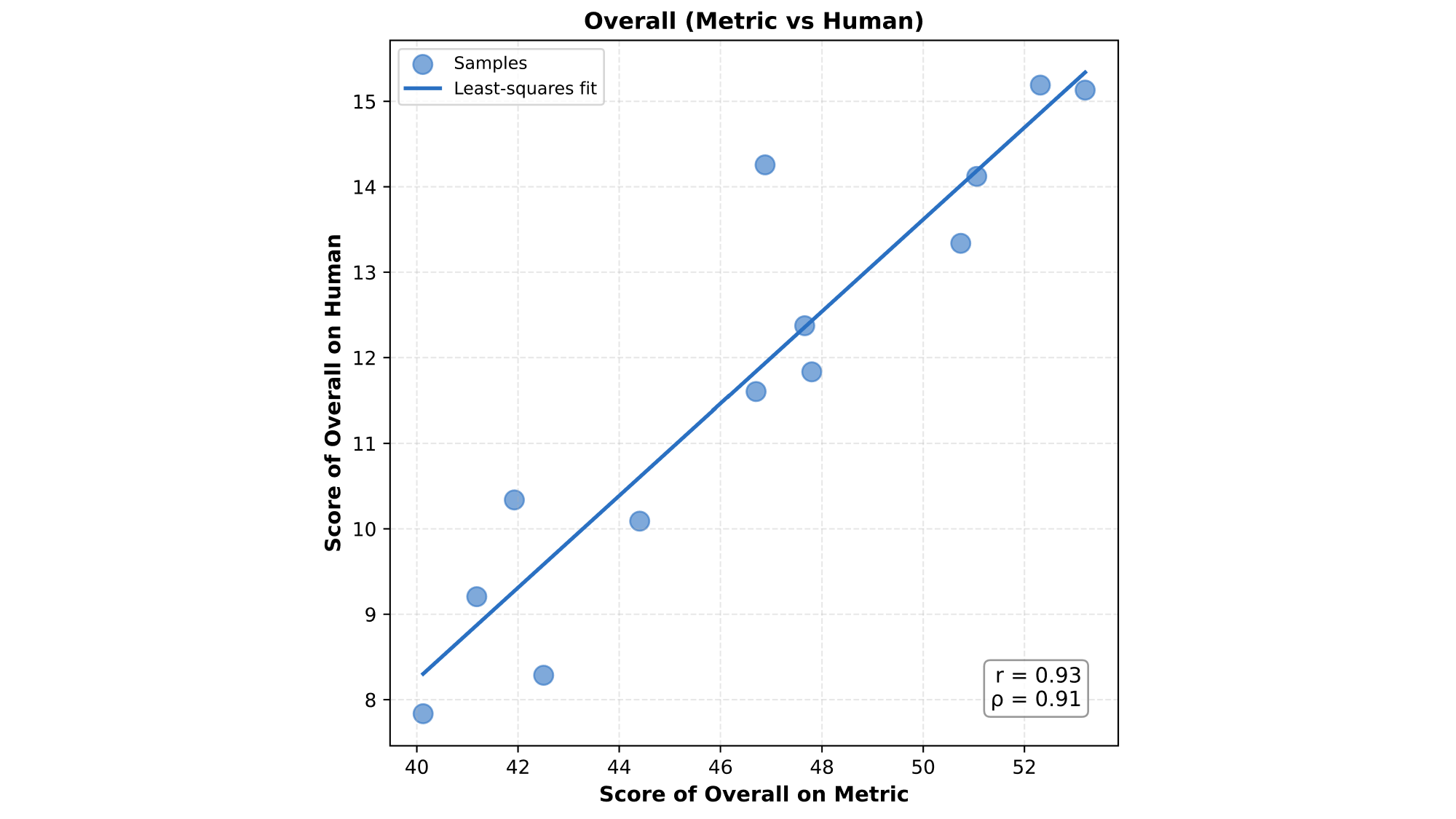}
    \caption{Correlation between WoW-World-Eval Overall Score and Human Eval.}
    \label{fig:correlation}
  \end{subfigure}
  \caption{\textbf{Side-by-side comparison:} (a) qualitative results by different models; (b) display a metric–preference correlation.}
  \label{fig:qual_corr_both}
\end{figure*}

\subsection{Qualitative Evaluation and Human Evaluation}
\label{ssec:qualitative}
\textbf{Qualitative Evaluation.} Figure~\ref{fig:visualization1} presents a qualitative comparison across models on a representative manipulation instruction, revealing failure patterns consistent with our quantitative results. General video generators such as CogVideoX, Cosmos-Predict1, and Wan2.1 exhibit visual artifacts, unintended camera motion, and even hallucinated human arms, reflecting dataset biases toward human demonstrations.

In terms of instruction understanding and planning, Cosmos-Predict1, WoW-Cosmos1, and WoW-Cosmos2 correctly execute the multi-step task, whereas CogVideoX and Wan2.1 fail despite producing motion. Cosmos-Predict2 and WoW-Wan show physics violations such as object duplication or motion without force, while Kling remains static and ignores the task, demonstrating limited planning capability.

Overall, the videos generated by WoW-
Cosmos2 is more likely to be indistinguishable from the real world in all aspects. These results highlight the need to evaluate models across all four dimensions—Video Quality, Instruction Understanding, Physical Law, and Planning Reasoning—as each captures a distinct failure mode in embodied video generation. Additional qualitative examples are included in the Appendix~\ref{sec:more_visualizations}.




\textbf{Human Evaluation.} To assess the validity of our evaluation metric, we conduct a human study with 15 domain experts, who rated over 1,200 videos, both real and generated, along four core dimensions that mirror our benchmark and examine the alignment between \textbf{WoW-World-Eval} and human judgment. As shown in Figure~\ref{fig:correlation}, both Pearson(r) and Spearman ($\rho$) correlations are reported for the Overall Score and Human Eval Score.

Results show a strong consistency between metric-based and human evaluations. The Overall Score demonstrates very strong Pearson and Spearman correlation with human judgment (r = 0.93, $\rho$ = 0.91), confirming that our benchmark provides a reliable and effective human-aligned evaluation of video generation performance.

\subsection{Turing Test}
\label{ssec:turing}
To demonstrate that our benchmark can serve as a principled foundation for Turing-test evaluations—and that performance on the benchmark is strongly correlated with outcomes in such Turing tests—while also addressing two remaining questions about embodied world models, we design and implement two complementary Turing Tests: \textit{Human Turing Test} and \textit{Inverse Dynamic Model Turing Test}.

\begin{figure}[!h]
  \includegraphics[width=0.47\textwidth]{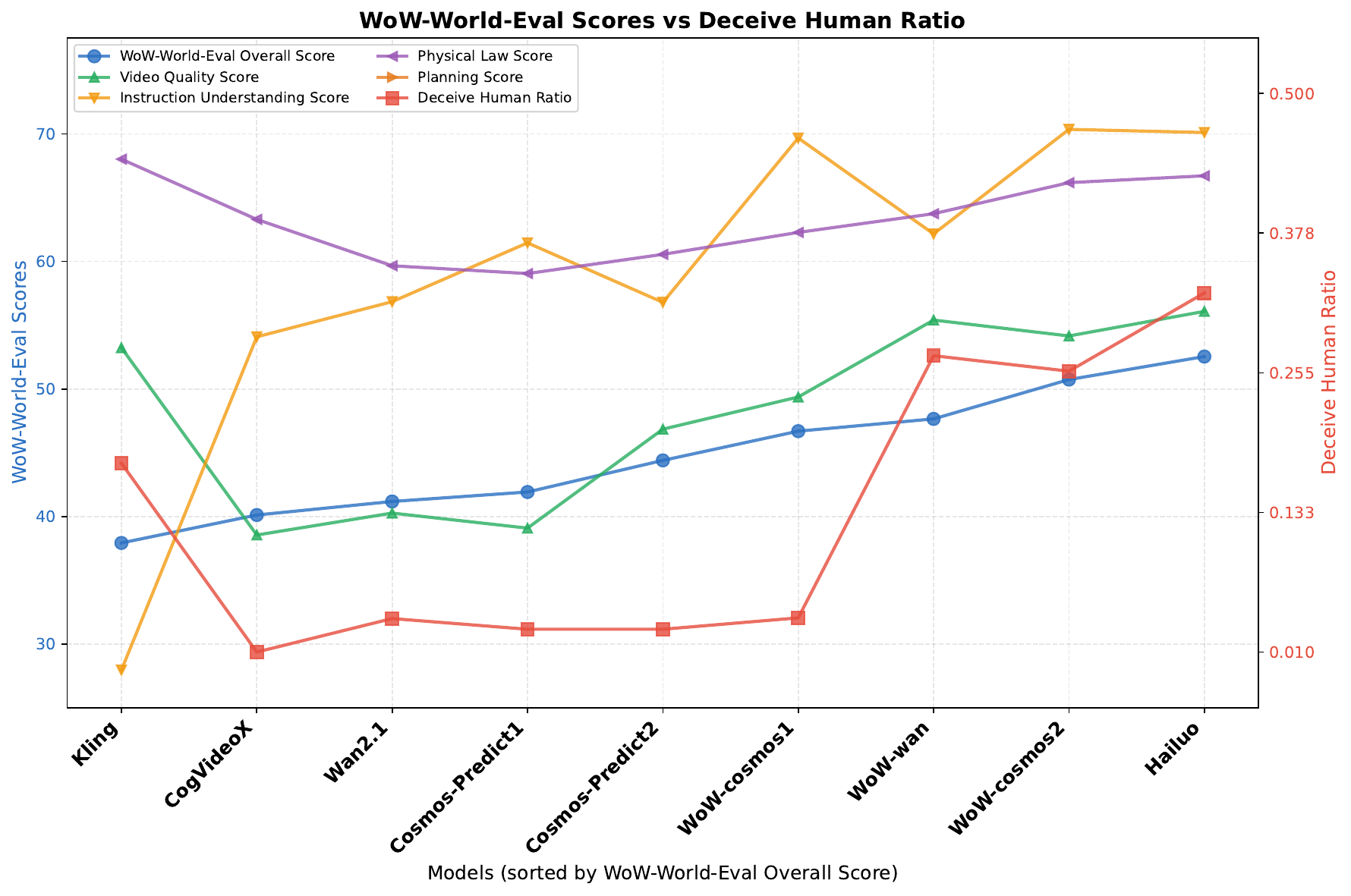} 
  \caption{\textbf{Trending between WoW-World-Eval Overall Score and Deceive Human Ratio.} }
  \label{fig:dual}
\end{figure}

\textbf{Human Turing Test.} Figure~\ref{fig:dual}  examines how \textbf{WoW-World-Eval} scores relate to the Deceive Human Ratio, measured by asking 13 participants to distinguish real videos from generated ones, which shows that models with higher overall scores consistently achieve higher deceive-human ratios, indicating stronger perceptual realism. For example, Hailuo and the WoW-cosmos2 variants are the most effective at fooling humans, consistent with their strong performance on our benchmark. This trend is statistically validated: the deceive-human ratio is strongly correlated with our Overall score (r = 0.679), meaning models that perform well on the benchmark are also more likely to fool humans. Breaking down by dimensions, Video Quality (r = 0.874) and Physical Law (r = 0.753) exhibit the highest correlations, showing that realism in both appearance and physical dynamics is what most convinces humans.

These results reveal a core insight: both visual quality and physical plausibility are essential for generated videos to appear real to humans. Videos lacking either fidelity or correct physics fail to deceive observers. This strong alignment with human perception underscores that quality and physics are indispensable components of an embodied world model benchmark.

\textbf{Inverse Dynamic Model Turing Test.}
We further introduce a Machine Turing Test using a real-world–trained Inverse Dynamics Model (IDM), GC-IDM~\cite{chi2025wow}, to evaluate whether generated videos exhibit physically executable dynamics. We collect 9 different manipulation tasks, from easy to hard, for evaluation. Details are in Appendix~\ref{ssec:IDM}

As shown in Table~\ref{tab:execution}, most high-scoring models on \textbf{WoW-World-Eval} still fail this test: Kling (9.88\%), Hailuo (2.47\%), and early open-source models all perform near 0\%, revealing that visual realism alone is insufficient for embodied execution. Models with real-robot data transfer significantly better—WoW-wan (40.74\%) and WoW-cosmos2 (18.52\%)—consistent with their strong Physical Law and Instruction Understanding scores.

These results highlight that: (i) The success rate of the generated videos on real robots is also correlated with our benchmark scores. (ii) Passing the IDM Turing Test requires both physics-grounded modeling and real-world exposure, emphasizing that true embodied competence goes beyond appearance quality. (iii) The videos generated by current embodied world models still have a significant gap from the real physical world.




\begin{table}[!h]
\centering
\caption{\textbf{Execution Accuracy on real world by Inverse Dynamic Model.}
All metrics: higher is better. Best results are \textbf{bold} with green highlight. Succ. Rate stands for Real-world success rate}
\label{tab:execution}
\setlength{\tabcolsep}{3pt} 
\begin{tabular}{lc}
\toprule
\textbf{Model} & \textbf{Real-world Succ. Rate (\%)} \\
\midrule
\textbf{Kling} & 9.88   \\
\textbf{Hailuo} & 2.47   \\
\midrule
\textbf{CogVideoX} & 0.00   \\
\textbf{Cosmos-Predict1} & 0.00 \\
\textbf{Wan2.1} & 0.00  \\
\textbf{Cosmos-Predict2} & 8.64   \\
\textbf{WoW-wan} & \cellcolor{mylightgreen}\textbf{40.74}   \\
\textbf{WoW-cosmos2} & 18.52  \\
\bottomrule
\end{tabular}
\end{table}

\section{Conclusion}
In this work, we present \textbf{WoW-World-Eval}, a comprehensive benchmark designed as a Turing Test for evaluating video foundation models in embodied AI. By combining fine-grained human preference assessments and the Inverse Dynamics Model (IDM), we provide a dual-perspective evaluation framework that examines both perceptual realism and physical executability. Through extensive experiments on 609 real-world robot manipulation data, we reveal that while some models achieve high perceptual fidelity, they struggle with long-horizon reasoning and real-world execution, often collapsing under the IDM Turing Test. These findings highlight a substantial gap between visual generation and embodied action, pointing to the urgent need for more physically grounded, generalizable world models for robotics. We hope our benchmark will serve as a foundation for future progress in building robust, general-purpose embodied intelligence.

\section{Acknowledge}
This work was supported by the National Natural Science Foundation of China (62476011), and by the Beijing Natural Science Foundation (L252060).
{
    \small
    \bibliographystyle{ieeenat_fullname}
    \bibliography{main}
}
\clearpage
\setcounter{page}{1}
\maketitlesupplementary

\section{More Related Works}
\label{sec:more_related_work}


\paragraph{Video Generation Models.}
The quest for scalable video synthesis has converged on two dominant architectures, both serving as backbones for predictive world modeling.

\textbf{Autoregressive Transformers.} 
Inspired by Large Language Models, architectures like VideoGPT~\cite{yan2021videogpt} treat video generation as a discrete sequence modeling task. By tokenizing frames via VQ-VAE~\cite{van2017neural}, these models predict visual token sequence-by-sequence. This paradigm, akin to next-word prediction, naturally aligns with the state-transition logic of world models. Notable examples like Genie~\cite{bruce2024genie} utilize this spatiotemporal autoregressive objective to build interactive environments and predictive models from diverse video data. However, the discrete quantization often imposes an upper bound on visual fidelity, resulting in the loss of high-frequency details.

\textbf{The Diffusion Paradigm.} 
Diffusion Models (DMs) have established a new state-of-the-art by formulating generation as iterative denoising. Early attempts like SVD~\cite{blattmann2023stable} adapted 2D U-Nets with temporal layers, excelling at texture but often treating time as a secondary dimension. The recent shift to Diffusion Transformers (DiT)~\cite{peebles2023scalable}, exemplified by Wan~\cite{wan2025wan}, Cosmos~\cite{cosmos2}, Sora~\cite{sora2024}, operates on unified spacetime patches. These models demonstrate emergent capabilities akin to a physics engine—simulating fluid dynamics and occlusion—making them ideal candidates for high-fidelity world simulation.

\paragraph{Embodied World Models.}
Unlike passive video generation, embodied world models function as predictive engines ($s_{t+1} | s_t, a_t$) to facilitate agent planning.

\textbf{From Latent States to Pixel Space.} 
Early research balanced between efficiency and detail. The Dreamer series~\citep{hafner2023mastering} prioritized computational tractability by learning dynamics in a compact latent space, enabling efficient RL planning but producing schematic visual reconstructions. Conversely, pixel-space approaches like RoboNet~\citep{dasari2019robonet} predicted future frames directly from interaction logs. While capturing texture, these models often struggled with the stochastic nature of real-world physics, leading to blurry predictions over long horizons.

\textbf{Generalist Simulators.} 
Recent efforts converge generative AI with robotics to create "Generalist Simulators." Foundation models like UniSim~\citep{yang2023unisim} and GAIA-1~\citep{hu2023gaia} leverage internet-scale data to learn high-level physics for zero-shot generalization. 

\section{Models Detail}
Since the models have different versions and parameter sizes, and also generate videos of different durations and resolutions, we introduce the models we use, their specific versions and parameter counts, as well as the generated durations and resolutions, to facilitate comparison and reproduction.
\paragraph{Kling.~\cite{Kling}}
A large-scale diffusion-transformer video generator known for photorealistic long-form outputs, strong motion control, and realistic physics. It employs a latent-space DiT architecture with 3D VAE compression and is optimized for high-consistency content such as sports or human motion. We used the latest Klingv2.1 to generate a 5-second 720p video.

\paragraph{Hailuo.~\cite{Hailuo}}
Hailuo is a commercial text-to-video and image-to-video model designed for high-quality short-form generation (6–10 seconds, up to 1080p). It emphasizes photorealism, smooth motion, and subject consistency, making it a strong closed-source baseline for perceptual quality. We use the publicly accessible Hailuo-02 for 5-second 768p video generation and evaluation.

\paragraph{CogVideoX.~\cite{yang2024cogvideox}}
An open diffusion-transformer video generator available in 2B/5B variants. It can produce ~10-second clips with improved motion expressiveness using a 3D causal VAE and MoE-style transformer backbone. The model supports both text-to-video and image-to-video generation and serves as a strong open-source baseline for long-duration video synthesis. For evaluation, we use CogVideoX-I2V-5B to generate 5-second 1360x768 videos.

\paragraph{Cosmos-Predict.~\cite{agarwal2025cosmos,cosmos2}}
NVIDIA’s open-source world-model family for video-based state prediction. It supports Text-/Image-/Video-to-World forecasting and can generate long-horizon, physics-aware video rollouts. The model is trained on large-scale real-world videos and emphasizes dynamics, object permanence, and 3D geometry consistency. We evaluate both Cosmos-Predict1 and the stronger Cosmos-Predict2 variants. Both Predict1\&2 are 5-second 720p, but the parameters of the model are 7B and 2B, respectively.

\paragraph{Wan.~\cite{wan2025wan}}
Wan is a series of open-source Diffusion Transformer video models (e.g., Wan 2.1/2.2) trained on large-scale image–video corpora. They support text-to-video and image-to-video generation, producing 5–10 second clips with strong spatial–temporal coherence. Wan models use 3D VAE compression and DiT backbones to improve motion consistency and are widely used as high-quality open video generation baselines. We fix the resolution for a 5-second 832x480 generation with Wan2.1-I2V-14B.

\paragraph{WoW.~\cite{chi2025wow}}
WoW is a 14B-parameter open-source embodied world model designed for physically grounded video prediction in robotic manipulation. It generates future video rollouts conditioned on visual observations and task instructions, with inductive biases (e.g., DINOv2 token distillation) that encourage spatial and physical consistency. Unlike generic video models, WoW explicitly targets planning, contact dynamics, and multi-step task execution. We use the different backbones of WoW, including Wan, Cosmos-Predict1, and Cosmos-Predict2, for evaluation; all generated video settings are the same as the backbone models.

\section{Detailed Metrics}
\label{sec:more_metrics}
\begin{figure*}[!h]
  \centering
  \includegraphics[width=1.0\textwidth]{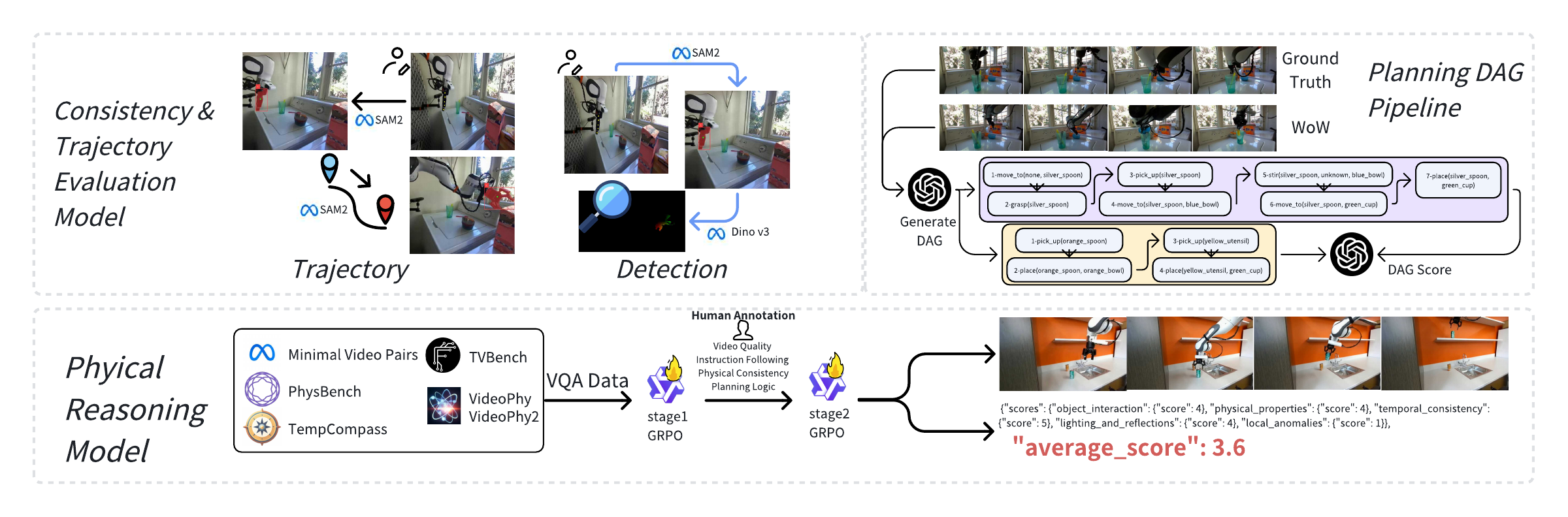} 
  \caption{\textbf{Overview of WoW-World-Eval Metrics.} }
  \label{fig:metric}
\end{figure*}

\subsection{Visual Fidelity.}
To evaluate the visual fidelity of generated videos, we adopt a suite of complementary metrics that capture low-level reconstruction accuracy, perceptual similarity, and distributional realism. These include 
FVD, SSIM, PSNR, DINO, and DreamSim.

\paragraph{Fréchet Video Distance (FVD).}
FVD is a distribution-level metric that measures how closely generated video features match those of real videos. 
Given feature means and covariances $(\mu_r, \Sigma_r)$ for real videos and $(\mu_g, \Sigma_g)$ for generated videos extracted using an I3D network, FVD is defined as:
\[
\mathrm{FVD}
= \|\mu_r - \mu_g\|_2^2
+ \mathrm{Tr}\!\left(\Sigma_r + \Sigma_g 
- 2\left(\Sigma_r\Sigma_g\right)^{1/2}\right).
\]
Lower FVD indicates higher spatial-temporal coherence and distributional realism.

\paragraph{Structural Similarity Index measures (SSIM).}
The Structural Similarity Index (SSIM) measures perceptual degradation in luminance, contrast, and structural information. 
For two frames $x$ and $y$:
\[
\mathrm{SSIM}(x,y) =
\frac{(2\mu_x\mu_y + C_1)(2\sigma_{xy} + C_2)}
     {(\mu_x^2 + \mu_y^2 + C_1)
      (\sigma_x^2 + \sigma_y^2 + C_2)},
\]
where $\mu$ and $\sigma$ denote means and variances of local patches. Higher SSIM corresponds to better perceptual consistency.

\paragraph{Peak signal-to-noise ratio (PSNR).}
PSNR evaluates pixel-level fidelity via the mean squared error (MSE) between reference and generated frames:
\[
\mathrm{PSNR}(x,y)
= 10\log_{10}\!\left(\frac{MAX^2}{\mathrm{MSE}(x,y)}\right).
\]
Higher PSNR generally indicates better reconstruction, although it is less aligned with human perception than SSIM or embedding-based metrics.

\paragraph{DINO Score.}
DINOv2 is a self-supervised visual foundation model trained with large-scale teacher--student distillation. 
We compute the cosine similarity between DINOv2 embeddings of generated and reference frames:
\[
\mathrm{DINO}(g_t, r_t)
= \frac{\langle f(g_t), f(r_t) \rangle}
       {\|f(g_t)\|_2 \, \|f(r_t)\|_2},
\]
where $f(\cdot)$ denotes the DINOv2 encoder. Higher similarity indicates closer semantic and structural alignment.

\paragraph{DreamSim.}
DreamSim combines CLIP, OpenCLIP, and DINO embeddings and is fine-tuned on human perceptual judgments to produce similarity scores aligned with human perception. For frames $x$ and $y$:
\[
\mathrm{DreamSim}(x,y) 
= 1 - \| E(x) - E(y) \|_2,
\]
where $E(\cdot)$ is the fused embedding.

\subsection{Instruction Semantic Correctness.}
These three scores—Caption Score, Sequence Match Score, and Execution Quality Score—allow us to evaluate instruction following in both supervised (with ground truth) and fully OOD settings. Empirically, all three metrics show strong agreement with human preference judgments, validating the reliability of our evaluation protocol.

\paragraph{Caption Score.}
For samples paired with ground-truth videos, we obtain a structured semantic representation of each video.
Using a templated prompt, GPT extracts a standardized caption containing: initial state, intermediate process, final state, object descriptions, and action descriptions.
The same procedure is applied to the model-generated video.
A vision–language model (VLM) compares the two captions along these five predefined components, assigning a score of 1 (match), 0.5 (partial match), or 0 (mismatch).
The Caption Score is computed as the mean over the five components.

\paragraph{Sequence Match Score.}
In this score, we directly analyze the alignment between the instruction and the actions depicted in the generated video.
GPT identifies the sequence of action–object pairs present in the video and aligns them with the intended execution order specified by the instruction.
We measure the degree to which the extracted sequence preserves the correctness of both action selection and temporal ordering, again using a discrete scale of 1 (match), 0.5 (partial match), or 0 (mismatch).
This yields the Sequence Match Score.

\paragraph{Execution Quality Score.}
To further evaluate execution fidelity, we assess how completely the depicted actions are carried out.
We employ a five-level rubric measuring the extent of object motion and action completion—from no meaningful change to full task completion.
\begin{enumerate}
    \item object does not move, and the action is not executed
    \item object slightly moves, or the action is partially executed
    \item object slightly moves, and the action is partially executed
    \item object reaches goal or action is fully executed
    \item object reaches goal and action is fully executed
\end{enumerate}
The score ranges from 1 to 5, producing the Execution Quality Score.
This score captures the degree to which the video evidences a coherent and physically plausible execution of the instructed task.

\subsection{Mask-guided Regional Consistency.}
\paragraph{Setup.} 
We evaluate the mask-guided regional consistency of each model based on three core components, as illustrated in Figure~\ref{fig:metric}:
(1) \textit{GroundingSAM-2} for point-guided segmentation and multi-frame tracking;  
(2) \textit{DINOv3-Large (standard)} as a frozen visual encoder to extract patch-level features;  
(3) \textit{Run-length Encoding (RLE)} for storing region masks efficiently across all frames.  
\paragraph{Implementation.}
We begin by manually identifying the first frame in which both the manipulated object and the robot gripper are clearly visible.  
A human annotator then places \textit{3--5 boundary points} along the visible contour of each region (object and gripper).  
These point sets provide fine-grained localization cues and are passed to GroundingSAM-2, which returns segmentation masks for the first frame and tracks them throughout the video. For each frame $t$, GroundingSAM-2 outputs an RLE mask for the object and an RLE mask for the gripper.  
If a region cannot be tracked in frame $t$ (due to occlusion or motion blur), we record its mask as an \textit{all-zero RLE mask}.  The background region is defined as the complement of the union of the two tracked regions:
\[
M^{\mathrm{bg}}_t = \mathbf{1} - \left(M^{\mathrm{obj}}_t \lor M^{\mathrm{arm}}_t\right).
\]
\noindent
Each RGB frame $I_t$ is encoded by DINOv3-Large to obtain a grid of patch features  
\[
F_t \in \mathbb{R}^{H_p \times W_p \times d}.
\]
All masks are decoded from RLE and downsampled to $(H_p, W_p)$ so they can serve as per-patch weights.  
For each region $r \in \{\mathrm{obj},\mathrm{arm},\mathrm{bg}\}$ we compute a normalized mask:
\[
w^\mathrm{r}_t(i,j) = 
\frac{M^\mathrm{r}_t(i,j)}{\sum_{i',j'} M^\mathrm{r}_t(i',j') + \epsilon},
\]
which becomes all zero when the region is missing.
\noindent
A region feature is obtained via mask-weighted averaging:
\[
\mathbf{f}^\mathrm{r}_t = \sum_{i,j} w^\mathrm{r}_t(i,j)\, F_t(i,j,:),
\]
followed by $\ell_2$ normalization if $\mathbf{f}^\mathrm{r}_t \neq 0$.

\paragraph{Consistency Score.}
We compute temporal consistency for each region separately using cosine similarity.  
Given normalized region features $\tilde{\mathbf{f}}^\mathrm{r}_t$, the consistency between two frames $a$ and $b$ for region $r$ is
\[
\mathrm{Consist}^\mathrm{r}(a,b) =
\begin{cases}
\tilde{\mathbf{f}}^\mathrm{r}_a \cdot \tilde{\mathbf{f}}^\mathrm{r}_b, & 
\|\tilde{\mathbf{f}}^\mathrm{r}_a\|_2 > 0,\, \|\tilde{\mathbf{f}}^\mathrm{r}_b\|_2 > 0, \\
0, & \text{otherwise.}
\end{cases}
\]
\noindent
For each frame $t \ge 2$ and region $r \in \{\mathrm{obj},\mathrm{arm},\mathrm{bg}\}$, we combine long-range and short-range temporal coherence:
\[
s^\mathrm{r}_t 
= \frac{1}{2}\,\mathrm{Consist}^\mathrm{r}(1,t)
+ \frac{1}{2}\,\mathrm{Consist}^\mathrm{r}(t-1,t).
\]
\noindent
The video-level Mask-guided Regional Consistency for region $r$ is then
\[
\mathrm{MRC}^\mathrm{r} = \frac{1}{T-1} \sum_{t=2}^{T} s^\mathrm{r}_t,
\]
where $T$ is the number of frames.  
In practice, we report three separate scores
\[
\bigl(\mathrm{MRC}^\mathrm{obj},\; \mathrm{MRC}^\mathrm{arm},\; \mathrm{MRC}^\mathrm{bg}\bigr),
\]
corresponding to the manipulated object, the gripper, and the background, respectively.

\subsection{Trajectory-level Consistency.}
We evaluate the trajectory-level consistency of each model by comparing the 2D motion trajectories of both the robot end-effector and manipulated objects in the generated videos against those in ground-truth (GT) videos. Human annotators first label representative keypoints on the robot arm and objects in the first frame; these serve as positive point prompts for SAM2 tracking, yielding dense and stable point tracks throughout the sequence, as illustrated in Figure~\ref{fig:metric}. The per-frame trajectory is defined as the mean of the tracked keypoints, normalized to the range $[0,1]^2$. We align the number of frames of the generated video and the GT video to the smaller one by uniformly sampling. We then compute three complementary metrics—L2Norm Error, Dynamic Time Warping (DTW), and Fréchet Distance—for both the robot end-effector and the manipulated object.

\paragraph{Keypoint labeling and SAM2 tracking.}
Given a video with frames $\{I_t\}_{t=1}^{T}$, human annotators place $N$ representative points on the robot end-effector or an object in the first frame:
\[
    \mathcal{K}_1 = \left\{
        \mathbf{u}_k^{(1)} \in \mathbb{R}^2
    \right\}_{k=1}^{N}, 
    \qquad 
    \mathbf{u}_k^{(1)} = (x_k^{(1)}, y_k^{(1)}).
\]
\paragraph{Keypoint prompting and SAM2 mask tracking.}
Given a video with frames $\{I_t\}_{t=1}^{T}$, human annotators place $N$ representative points on the robot end-effector or an object in the first frame. These points serve as positive prompts for SAM2, which returns a binary segmentation mask for the object at each frame:
\[
    \mathbf{M}^{(t)} \in \{0,1\}^{H \times W}, 
    \qquad t = 1,\dots,T.
\]
Each mask $\mathbf{M}^{(t)}$ defines the set of foreground pixels
\[
    \Omega^{(t)} = 
    \big\{ (x,y) \;\big|\; \mathbf{M}^{(t)}(x,y)=1 \big\}.
\]

\paragraph{Mask-to-point trajectory conversion.}
To obtain a 2D trajectory from the mask sequence, we compute the centroid of the foreground region:
\[
    \mathbf{p}_t
    = (x_t, y_t) =
    \frac{1}{|\Omega^{(t)}|}
    \sum_{(x,y)\in\Omega^{(t)}} 
    (x,y).
\]
Let the video resolution be $(W,H)$. We then normalize all coordinates to $[0,1]^2$:
\[
    \hat{\mathbf{p}}_t
    =
    \left(
        \frac{x_t}{W},
        \frac{y_t}{H}
    \right),
    \qquad t = 1,\dots,T.
\]
This centroid simplification provides a stable 2D trajectory independent of resolution for each tracked component.


\paragraph{Camera-motion–aware trajectory correction.}
In generated videos with unstable viewpoints, the observed end-effector trajectory entangles true robot motion with camera drift or jitter, leading to biased evaluations. To recover the actual end-effector motion, we subtract the estimated camera trajectory $\mathbf{c}_t$ from the observed end-effector trajectory $\hat{\mathbf{p}}_t$:
\[
    \mathbf{p}_t^{\mathrm{true}}
    =
    \hat{\mathbf{p}}_t - \mathbf{c}_t.
\]
This correction yields a cleaner and more comparable motion signal across videos. Empirically, smaller camera metrics ATE/RPE correspond to more reliable corrected trajectories, whereas larger camera errors introduce evaluation noise and indicate the need for improved video generation or stabilization.

\paragraph{Camera trajectory estimation.}
   Given a video, we uniformly sample frames $\{I_t\}_{t=1}^{T}$ and detect sparse Shi--Tomasi keypoints on the image boundary of the first frame 
   \[ \mathcal{P}_1 = \big\{\mathbf{u}_k^{(1)} \in \mathbb{R}^2\big\}_{k=1}^{N}, \qquad \mathbf{u}_k^{(1)} = (x_k^{(1)}, y_k^{(1)}), 
   \]
    where boundary regions are assumed to be dominated by a static background. We then track these keypoints across subsequent frames with pyramidal Lucas--Kanade optical flow, obtaining correspondences between a reference frame and frame $t$:
    \[
        \mathcal{P}_{\text{ref}} = \{\mathbf{u}_k^{\text{ref}}\}, \qquad
        \mathcal{P}_t = \{\mathbf{u}_k^{(t)}\}.
    \]
    For each frame $t$, we robustly estimate a 2D affine transform (restricted to rotation/scale/translation) via RANSAC:
    \[
        \mathbf{A}_t =
        \begin{bmatrix}
            a_{11} & a_{12} & t_x \\
            a_{21} & a_{22} & t_y
        \end{bmatrix}, \quad
        \mathbf{u}_k^{(t)} \approx \mathbf{A}_t
        \begin{bmatrix}
            \mathbf{u}_k^{\text{ref}} \\ 1
        \end{bmatrix},
    \]
    where $\mathbf{t}_t = (t_x, t_y)$ captures the apparent translation of the static background in the image plane. Because camera motion is opposite to background motion, we define the per-frame camera displacement as
    \[
        \Delta \mathbf{c}_t = -\,\mathbf{t}_t = (-t_x,\,-t_y).
    \]
    Starting from $\mathbf{c}_1 = (0,0)$, we accumulate these displacements over time, with an additional drift clipping to suppress implausibly large jumps between adjacent frames:
    \[
        \mathbf{c}_t = \mathbf{c}_{t-1} + \Delta \mathbf{c}_t, \quad t = 2,\dots,T,
    \]
    where $\mathbf{c}_t = (x_t, y_t)$ denotes the cumulative camera offset (in pixels) at frame $t$ relative to the first frame. The original frame resolution $(W, H)$ is stored together with the trajectory.


\paragraph{Temporal alignment and normalization across videos.}
For a GT video and a generated video, we obtain two normalized trajectories:
\[
    \{\hat{\mathbf{p}}_t^{\text{gt}}\}_{t=1}^{T_{\text{gt}}}, \qquad
    \{\hat{\mathbf{p}}_t^{\text{gen}}\}_{t=1}^{T_{\text{gen}}}.
\]
To compare two trajectories of different lengths, we first determine the target number of frames:
\[
    T = \min(T_{\text{gt}}, T_{\text{gen}}).
\]
We uniformly sample the longer trajectory to obtain exactly \(T\) frames.  
Let the aligned trajectories be
\[
    \mathbf{q}_t^{\text{gt}} = \hat{\mathbf{p}}_{s^1(t)}^{\text{gt}}, 
    \qquad
    \mathbf{q}_t^{\text{gen}} = \hat{\mathbf{p}}_{s^2(t)}^{\text{gen}}, 
    \qquad 
    t = 1,\dots,T.
\]
where $s^1$ and $s^2$ are the sampled indexes.

\paragraph{Absolute Trajectory Error (ATE).}
    ATE measures the absolute discrepancy between the generated and GT camera trajectories over the entire sequence. We define
    \[
        \mathrm{ATE}
        =
        \sqrt{
            \frac{1}{T}
            \sum_{t=1}^{T}
            \big\|
                \hat{\mathbf{c}}_t^{\text{gen}} - \hat{\mathbf{c}}_t^{\text{gt}}
            \big\|_2^{2}
        },
    \]
    where $\|\cdot\|_2$ denotes the L2 norm. A lower ATE indicates that the overall camera path in the generated video more closely follows the ground-truth trajectory in the image plane.

    \paragraph{Relative Pose Error (RPE).}
    To evaluate the fidelity of \emph{local} camera motion (e.g., instantaneous velocity and direction), we compute the Relative Pose Error. For a temporal offset $\Delta$ (we use $\Delta = 1$ frame in our experiments), we define normalized relative motion as
    \[
        \begin{aligned}
            \mathbf{v}_t^{\text{gt}}
            &=
            \hat{\mathbf{c}}_{t+\Delta}^{\text{gt}} - \hat{\mathbf{c}}_t^{\text{gt}}, \qquad
            \mathbf{v}_t^{\text{gen}}
            =
            \hat{\mathbf{c}}_{t+\Delta}^{\text{gen}} - \hat{\mathbf{c}}_t^{\text{gen}},
            \\[4pt]
            &\qquad\qquad t = 1,\dots,T-\Delta.
        \end{aligned}
    \]
    The RPE is then defined as
    \[
        \mathrm{RPE}
        =
        \sqrt{
            \frac{1}{T - \Delta}
            \sum_{t=1}^{T-\Delta}
            \big\|
                \mathbf{v}_t^{\text{gen}} - \mathbf{v}_t^{\text{gt}}
            \big\|_2^{2}
        }.
    \]
    RPE focuses on frame-to-frame motion consistency: lower values indicate that the generated video better matches the ground-truth in terms of local camera dynamics (e.g., smoothness, acceleration patterns, and motion direction).

\paragraph{L2Norm Error.}
The per-frame L2Norm error measures instantaneous positional discrepancy:
\[
    \mathrm{L2norm}
    =
    \sqrt{
        \frac{1}{T}
        \sum_{t=1}^{T}
        \big\|
            \mathbf{q}_t^{\text{gen}}
            -
            \mathbf{q}_t^{\text{gt}}
        \big\|_2^2
    }.
\]
A lower ED indicates that the generated trajectory more closely follows the GT path frame by frame.

\paragraph{Dynamic Time Warping (DTW).}
DTW preserves the shape similarity of two trajectories while allowing non-linear temporal alignment. The DTW cost is defined as
\[
    \mathrm{DTW}(\mathbf{q}^{\text{gen}}, \mathbf{q}^{\text{gt}})
    =
    \min_{\pi}
    \sum_{(t, s)\in \pi}
    \big\|
        \mathbf{q}_t^{\text{gen}}
        -
        \mathbf{q}_s^{\text{gt}}
    \big\|_2,
\]
where $\pi$ is a monotonic warping path. Lower DTW values indicate higher similarity in trajectory geometry, regardless of speed variations.

\paragraph{Fréchet Distance.}
The Fréchet Distance captures the minimum leash-length needed for two curves to be traversed in order, providing a strict measure of geometric similarity:
\[
    \mathrm{FD}(\mathbf{q}^{\text{gen}}, \mathbf{q}^{\text{gt}})
    =
    \inf_{\alpha,\beta}
    \max_{t \in [0,1]}
    \left\|
        \mathbf{q}^{\text{gen}}_{\alpha(t)}
        -
        \mathbf{q}^{\text{gt}}_{\beta(t)}
    \right\|_2,
\]
where $\alpha$ and $\beta$ are continuous, non-decreasing reparameterizations. Unlike DTW, the Fréchet metric enforces simultaneous forward progression along the curves.

\subsection{Physical and Causal Reasoning.}

    \subsubsection*{Base Model and Optimization Algorithm}
To implement automated evaluation, we selected \textbf{Qwen-2.5-VL (7B)}~\cite{bai2025qwen25vltechnicalreport} as our base vision-language model. We employed the \textbf{GRPO} algorithm, optimizing the model through a two-stage fine-tuning process to equip it with precise physical commonsense understanding and scoring capabilities, as illustrated in Figure~\ref{fig:metric}.

\subsubsection*{The GRPO Algorithm}
GRPO optimizes the policy model $\pi_\theta$ by maximizing an objective that encourages high-reward outputs while maintaining stability via a clipping mechanism and KL-divergence regularization.

For each input query $q$, GRPO samples a group of outputs $\{o_1, o_2, \cdots, o_G\}$ from the old policy $\pi_{\theta_{old}}$. The optimization objective is defined as:
\begin{equation}
\begin{aligned}
\mathcal{J}_{GRPO}(\theta) &= \mathbb{E}\Bigg[ \frac{1}{G} \sum_{i=1}^{G} \Bigg( \min \Bigg[ \\
&\quad \frac{\pi_\theta(o_i|q)}{\pi_{\theta_{old}}(o_i|q)} \hat{A}_i, \\
&\quad \text{clip}\left( \frac{\pi_\theta(o_i|q)}{\pi_{\theta_{old}}(o_i|q)}, 1-\varepsilon, 1+\varepsilon \right) \hat{A}_i \Bigg] \\
&\quad - \beta D_{KL}[\pi_\theta || \pi_{ref}] \Bigg) \Bigg]
\end{aligned}
\end{equation}
where $\varepsilon$ is the clipping hyperparameter, and $\beta$ controls the KL-divergence penalty with respect to the reference model $\pi_{ref}$.

The advantage $\hat{A}_i$ for the $i$-th output is calculated based on the relative rewards within the sampled group. Specifically, assuming output $o_i$ receives a reward score $r_i$, the advantage is computed by standardizing the rewards:
\begin{equation}
\hat{A}_i = \frac{r_i - \text{mean}(\{r_1, \dots, r_G\})}{\text{std}(\{r_1, \dots, r_G\}) + \epsilon}
\end{equation}
This group-relative formulation serves as a dynamic baseline, effectively reducing variance without requiring a separate value network.

\subsubsection*{Stage 1: Foundational Video Understanding Fine-tuning}
\textbf{Objective:} To imbue the model with a foundational understanding of physical events and causal relationships.
\textbf{Data and Method:} We utilized a dataset of approximately 50,000 samples from six video understanding benchmarks. All data was formatted as multiple-choice Video Question Answering (VQA) tasks.
\begin{itemize}
    \item \textbf{Sampling:} For each video-question pair $q$, we sampled a group of outputs of size $G=8$.
    \item \textbf{Reward Calculation:} We implemented a rule-based reward function. If a sampled output $o_i$ correctly matched the ground-truth option (e.g., "Option A"), it received a reward $r_i = 1.0$; otherwise, $r_i = 0.0$.
    \item \textbf{Optimization:} The model was updated using the calculated group advantages to increase the probability of generating correct answers.
\end{itemize}

\textbf{Results:} Stage 1 fine-tuning significantly improved the model's video comprehension abilities. On a comprehensive held-out test set from these benchmarks, the model's average accuracy increased from \textbf{60.83\%} (the Qwen-2.5-VL 7B backbone) to \textbf{71.51\%}.

\subsubsection*{Stage 2: Scoring Alignment Fine-tuning}
\textbf{Objective:} To train the model to follow instructions and output quantitative 1-5 scores for four dimensions in a JSON format, aligning with human-annotated standards.
\textbf{Method:} We continued to use the GRPO algorithm, fine-tuning on our internally annotated dataset of 1,297 data points. We used the following prompt template to guide the model in generating the scoring JSON.
\textbf{Reward Function:}
In Stage 2, the GRPO optimization objective was to maximize the expectation of a reward function $R(y_{out}, y_{gt})$, which quantifies the alignment between the model-generated JSON $y_{out}$ and the ground-truth human-annotated JSON $y_{gt}$.

The reward $R$ is calculated as follows:
\begin{enumerate}
    \item Parse the JSON objects from $y_{out}$ and $y_{gt}$.
    \item Let $\mathcal{K}$ be the set of four evaluation keys (i.e., \texttt{video\_quality}, \texttt{instruction\_following}, \texttt{physical\_consistency}, \texttt{planning\_logic}).
    \item For each key $k \in \mathcal{K}$ present in $y_{gt}$, if $k$ is also present in $y_{out}$, calculate the normalized absolute error $e_k$.
    \item Before calculating $e_k$, the scores $s_k^{gt}$ from $y_{gt}$ and $s_k^{out}$ from $y_{out}$ are clipped to the valid range $[1.0, 5.0]$, yielding $s_k'^{gt}$ and $s_k'^{out}$.
    \item The normalized absolute error $e_k$ is defined as:
    \begin{equation}
    e_k = \frac{|s_k'^{gt} - s_k'^{out}|}{4.0}
    \end{equation}
    The denominator 4.0 (i.e., $5-1$) normalizes the error to the range $[0.0, 1.0]$.
    \item The mean error $\bar{e}$ is calculated over all matched keys $\mathcal{K}_{match} \subseteq \mathcal{K}$:
    \begin{equation}
    \bar{e} = \frac{1}{|\mathcal{K}_{match}|} \sum_{k \in \mathcal{K}_{match}} e_k
    \end{equation}
    \item The final reward $R$ is defined as $1.0 - \bar{e}$ and clipped to $[0.0, 1.0]$:
    \begin{equation}
    R(y_{out}, y_{gt}) = \max(0.0, \min(1.0, 1.0 - \bar{e}))
    \end{equation}
    If $y_{out}$ or $y_{gt}$ are not valid JSON, or if $\mathcal{K}_{match}$ is empty, the reward $R=0.0$.
\end{enumerate}

\subsubsection*{Final Inference for Evaluation}
After completing the two-stage fine-tuning, we obtained an evaluation model with a strong understanding of physical common sense (from S1) and the ability to perform structured scoring tasks (from S2). For the final automated evaluation in our paper, we employed a detailed, zero-shot inference prompt. This prompt instructs the model to focus \textit{exclusively} on physical plausibility, refining its evaluation criteria into the six dimensions used in our main analysis.

\subsection{Overall Benchmark Score.}
We normalize all metrics to a unified three-digit scale of 0–100 to ensure consistent interpretation and cross-metric comparability. Metrics that naturally fall within this range are retained as-is. For the remaining metrics (including FVD and PSNR, we fix their boundaries up to 2000 and 50, respectively), we first map their values to [0,1], then apply a monotonic transformation, and finally scale the result to [0,100].

To determine the optimal mapping method and parameter for each metric, we use a subset of human evaluation scores as supervision. Specifically, we apply different values within a grid over [0,5], compute the Pearson correlation between the transformed metric scores and human ratings, and select the method and parameter that maximizes this correlation. 

\textbf{Parametric mappings.}
\label{ssec:mappings}
After pre-scaling, we apply a single-parameter monotone transform $f_m(\cdot;\theta_m)$ and then rescale to $(0,100)$:
\[
s_{i,m}=100\,f_m\!\big(\hat x_{i,m};\theta_m\big),\qquad s_{i,m}\in(0,100).
\]
We consider the following families (all are strictly increasing on $[0,1]$):
\[
\begin{aligned}
\textbf{Simple:} &\quad f_(x) = x,\\[3pt]
\textbf{Power (Gamma):} &\quad f_\gamma(x) = x^\gamma,\ \gamma>0,\\[3pt]
\textbf{Logit temperature:} &\quad
f_T(x) = \sigma(\mathrm{logit}(x)/T),\ T>0,\\
&\quad \sigma(t)=\tfrac{1}{1+e^{-t}},\\[3pt]
\textbf{Tanh slope:} &\quad
f_\kappa(x)=\tfrac12(\tanh(\kappa(2x-1))+1),\\ 
&\quad \kappa>0.
\end{aligned}
\]
In practice, $\gamma>1$ accentuates the high end, while $T<1$ or $\kappa>1$ expands the mid-range and compresses extremes.
For numerical stability with logit we use a small $\varepsilon$ (e.g., $10^{-6}$) and replace $x$ by $\mathrm{clip}(x;\varepsilon,1-\varepsilon)$ only inside $\mathrm{logit}(\cdot)$.

\textbf{Parameter selection and freezing.}
For each metric $m$, $\theta_m\in\{\gamma,T,\kappa\}$ is selected on a fixed development set by maximizing
the Fisher-$z$ averaged Pearson correlation between $f_m(\hat x;\theta)$ and human ratings across $K$-fold CV;
Spearman correlation is used as a tie-breaker. The chosen $\theta_m$ is then \emph{frozen} and applied to all evaluations.

The resulting mapping methods and $\theta_m$ values for all metrics $m$ are summarized in Table~\ref{tab:gamma-param}.
\begin{table}[!h]
\centering
\caption{\textbf{The mapping parameters for all the metrics.}}
\label{tab:gamma-param}
\begin{tabular}{lcc}
\toprule
\textbf{Metric} & \textbf{Mapping} & \textbf{Parameter} \\
\midrule
\textbf{FVD} & gamma & 1.52 \\
\textbf{PSNR} & tanh & 4.71 \\
\textbf{SSIM} & gamma & 0.61 \\
\textbf{DINO} & gamma & 3.06 \\
\textbf{DreamSim} & gamma & 2.94 \\
\textbf{Caption Score} & gamma & 0.12 \\
\textbf{Seq. Match Score} & gamma & 2.45 \\
\textbf{Exec. Qual Score} & gamma & 2.97 \\
\textbf{Planning DAG} & simple & - \\
\textbf{Robot Con.} & gamma & 2.93 \\
\textbf{Obj. Con.} & tanh & 4.93 \\
\textbf{Scene Con.} & gamma & 3.94 \\
\textbf{Robot Traj. L2norm} & gamma & 2.86 \\
\textbf{Robot Traj. DTW} & gamma & 1.87 \\
\textbf{Robot Traj. FD} & gamma & 4.00 \\
\textbf{Obj. Traj. L2norm} & gamma & 1.27 \\
\textbf{Obj. Traj. DTW} & gamma & 2.99 \\
\textbf{Obj. Traj. FD} & gamma & 3.52 \\
\textbf{Physical Score} & simple & - \\
\textbf{Cam. ATE} & simple & - \\
\textbf{Cam. RPE} & simple & - \\

\bottomrule
\end{tabular}
\end{table}

\section{Extended Experiment Analysis}
\subsection{WoW-World-Eval Human Evaluation Rule}
We evaluate generated videos along four independent dimensions: Video Quality, Instruction Understanding, Physical Law, and Planning.
Each dimension is scored on a 1–5 scale, and the Overall Score is defined as the sum of the four scores (range: 4–20).
All dimensions are assessed independently.

\paragraph{Video Quality.} This dimension assesses the perceptual integrity of the video itself, independent of task correctness or physical feasibility. The focus is strictly on the clarity, stability, and visibility of the content presented. Scoring Criteria will be as follows:

\textbf{5 — Excellent}\par
\begin{itemize}
    \item High spatial clarity with sharp and stable rendering
    \item Proper exposure and color balance; no notable artifacts or noise
    \item All key elements remain continuously visible
    \item No distracting visual irregularities
\end{itemize}

\textbf{4 — Good}\par
\begin{itemize}
    \item Minor imperfections (e.g., slight blur, mild exposure fluctuation, low-level noise)
    \item Overall visibility remains unaffected
\end{itemize}
 
\textbf{3 — Fair}\par
\begin{itemize}
    \item Multiple noticeable issues, including moderate blur, frequent autofocus/exposure changes, or significant noise
    \item Occasional hindrance to understanding the scene
    \item Overall content remains interpretable
\end{itemize}

\textbf{2 — Poor}\par
\begin{itemize}
    \item Significant visual defects: strong blur, over/under-exposure, instability, occlusion of key regions
    \item Compression artifacts or frame drops are obvious and disruptive
    \item Core task content is frequently hard to interpret
\end{itemize}

\textbf{1 — Unusable}\par
\begin{itemize}
    \item Severe degradation: extreme darkness/brightness, pervasive blur, heavy artifacts, or constant frame loss
    \item Critical task elements cannot be identified
    \item The video cannot support meaningful evaluation
\end{itemize}

\paragraph{Instruction Understanding.} This dimension measures whether the final outcome satisfies the given instruction.
It is strictly goal-oriented, disregarding process efficiency or motion plausibility.
Scoring Criteria will be as follows:

\textbf{5 — Fully Achieved}\par
\begin{itemize}
    \item The instruction is completed precisely and entirely
    \item All expected final conditions are satisfied
    \item No essential elements are missing
\end{itemize}

\textbf{4 — Mostly Achieved}\par
\begin{itemize}
    \item The majority of the instructions is correctly executed
    \item Minor deviations exist, but do not compromise overall task success
\end{itemize}
 
\textbf{3 — Partially Achieved}\par
\begin{itemize}
    \item Only some components of the instruction are fulfilled
    \item Significant omissions or inaccuracies remain
    \item The intended goal is only partially realized
\end{itemize}

\textbf{2 — Minimally Achieved}\par
\begin{itemize}
    \item Most of the required outcome is not completed
    \item The behavior exhibits weak or inconsistent correspondence to the instruction
\end{itemize}

\textbf{1 — Not Achieved}\par
\begin{itemize}
    \item The instruction is not fulfilled at all
    \item Actions are irrelevant or contradictory to the task
    \item A fully static video always receives a score of 1 in this dimension
\end{itemize}

\paragraph{Physical Law.} This dimension evaluates whether the depicted motions and object interactions adhere to real-world physical principles, including dynamics, continuity, and plausible human-robot interaction.
Scoring Criteria will be as follows:

\textbf{5 — Fully Physical}\par
\begin{itemize}
    \item All motions are dynamically coherent and physically plausible
    \item Object interactions follow realistic physical responses
    \item No penetration artifacts or implausible transitions
    \item A completely static video (identical to the first frame) is also scored as 5, as it contains no physical violations
\end{itemize}

\textbf{4 — Mostly Physical}\par
\begin{itemize}
    \item Minor deviations from realistic physics, but overall behavior remains plausible
    \item No significant physical inconsistencies
\end{itemize}
 
\textbf{3 — Moderately Physical}\par
\begin{itemize}
    \item Noticeable but non-catastrophic physical inconsistencies
    \item Partial deviations from expected object dynamics
    \item Movements occasionally appear unnatural
\end{itemize}

\textbf{2 — Poorly Physical}\par
\begin{itemize}
    \item Frequent or severe physics violations
    \item Object motion or robot kinematics contradict basic real-world principles
    \item Motion discontinuities or penetrations are common
\end{itemize}

\textbf{1 — Physically Impossible}\par
\begin{itemize}
    \item Major violations of fundamental physics (e.g., teleportation, reverse gravity, object behavior without force causality)
    \item Substantial geometric inconsistencies (large penetrations)
    \item If a generated human performs part or all of the task, the Physical Law score is supposed to be 1, regardless of physical plausibility
\end{itemize}

\paragraph{Planning Reasoning.} This dimension assesses the logical structure and coherence of the robot’s action sequence, independent of final task success.
It captures whether the robot demonstrates purposeful, ordered planning.
Scoring Criteria will be as follows:

\textbf{5 — Well-Structured Planning}\par
\begin{itemize}
    \item Action sequence is coherent, efficient, and well aligned with the task
    \item No extraneous or aimless motions
    \item Exhibits deliberate, task-driven progression
\end{itemize}

\textbf{4 — Reasonable Planning}\par
\begin{itemize}
    \item Overall logical sequence with minor redundant actions
    \item Task progression remains clear and purposeful
\end{itemize}
 
\textbf{3 — Weak Planning}\par
\begin{itemize}
    \item Action sequence contains irregularities or inefficiencies
    \item Task intent is visible, but execution lacks structure
    \item Multiple corrective attempts may be present
\end{itemize}

\textbf{2 — Poor Planning}\par
\begin{itemize}
    \item Disorganized or inconsistent action sequence
    \item Repeated irrelevant behaviors or unnecessary back-and-forth motions
    \item Weak alignment between steps and task objective
\end{itemize}

\textbf{1 — No Planning}\par
\begin{itemize}
    \item Actions are random, chaotic, or entirely absent
    \item A fully static video always receives a score of 1 in this dimension
\end{itemize}

\paragraph{Overall Score.} 
The Overall Score is defined as the sum of the four scores. All dimensions are weighted equally, supporting transparent and reproducible evaluation.

\subsection{All correlation of WoW-World-Eval score and Human Preference}
\begin{figure*}[!t]
  \centering
  \includegraphics[width=1.0\textwidth]{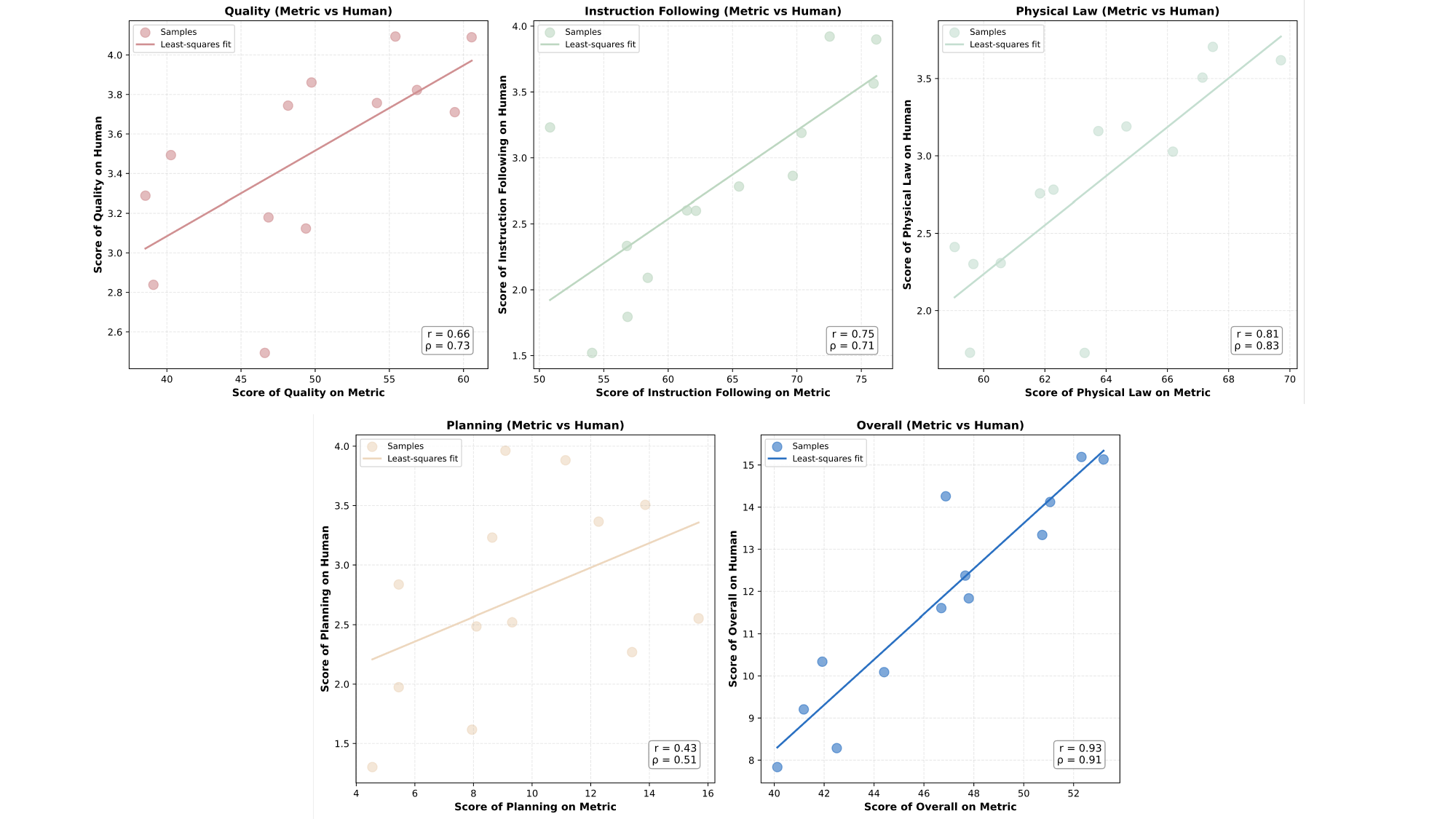} 
  \caption{\textbf{All correlation for WoW-World-Eval and Human Preference.} }
  \label{fig:all_correlation}
\end{figure*}

In section~\ref{ssec:quantitative}, we report the correlation between our overall scores and overall human preferences. However, to quantify how well WoW-World-Eval reflects human perception in all perspectives, we also provide the correlation between these four metrics—Video Quality, Instruction Understanding, Physical Law, and Planning. Figure~\ref{fig:all_correlation} reports both strong Pearson (r) and Spearman ($\rho$) correlations between all groups of benchmark scores and human preferences, which, importantly, confirms that metric-based evaluation can reliably approximate human judgment, supporting the use of WoW-World-Eval as a scalable alternative to costly human evaluations.

\paragraph{Video Quality.}
Quality metrics exhibit a quite strong correlation with human assessments (r = 0.66, $\rho$ = 0.73). This confirms that perceptual sharpness, temporal consistency, and high-level structure—captured jointly by FVD, PSNR, SSIM, DINO, and DreamSim—align well with how humans evaluate visual sensitivity.

\paragraph{Instruction Understanding.}
The Instruction Understanding score shows robust alignment with human preferences (r = 0.75, $\rho$ = 0.71). Models that accurately capture state transitions, object–action relationships, and multi-step task semantics tend to be judged by humans as better at “following instructions,” validating our caption-, sequence- and execution-based IU metrics.

\paragraph{Physical Law.}
Correlation is highest for Physical Law (r = 0.81, $\rho$ = 0.83). This indicates that humans are highly sensitive to violations of physics—object irregular motion, penetration, trajectory discontinuities, implausible state changes—and that our physics metrics effectively capture such failures. 

\paragraph{Planning.}
Planning demonstrates only moderate correlation with human ratings (r = 0.43, $\rho$ = 0.51). This reflects the inherent difficulty of automatically evaluating long-horizon reasoning and structured action sequences, as well as the limited planning ability of current video models. 

\subsection{Ground-truth video replay of IDM}
\label{ssec:IDM}
To ensure that our Inverse Dynamics Model (IDM) provides reliable supervision in the IDM Turing Test, we first validate the accuracy of a reproduced Gripper-Centric IDM (GC-IDM) following WoW~\cite{chi2025wow}. Before applying the IDM to model-generated videos, it is essential to verify that its action inference is sufficiently accurate on ground-truth real-world data; otherwise, low execution accuracy could stem from model weakness rather than the generated video being physically unrealistic.

\begin{figure*}[!h]
  \centering
  \includegraphics[width=1.0\textwidth]{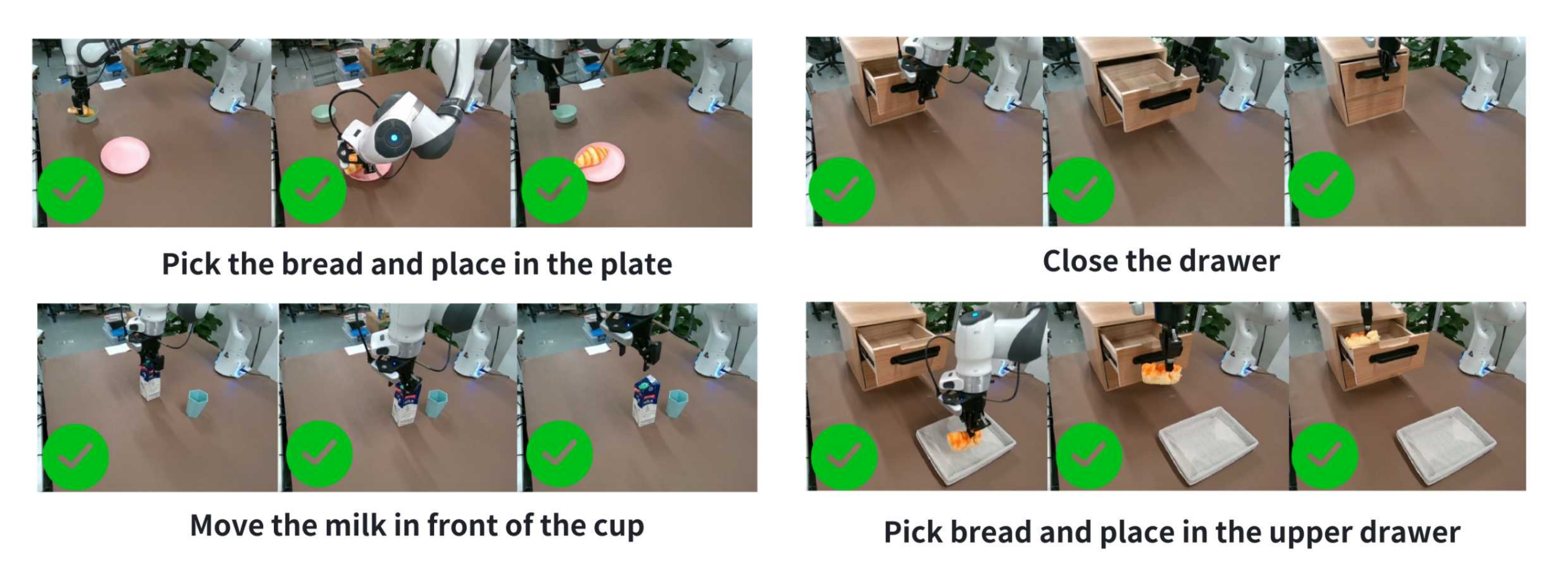} 
  \caption{\textbf{Real-World ground-truth video replay on real robot.} }
  \label{fig:real-world}
\end{figure*}

We therefore evaluate GC-IDM across 9 manipulation tasks, each with 10 ground-truth execution videos. They are Easy (Pick bread and place on the plate; Close the drawer; Move the milk in front of the cup), Medium (Pick bread and place it in the upper drawer; Take the cup off the cup holder; Open the drawer) and Hard (Flip the button to the correct position; Hang the cup on the cup holder; Insert chopsticks into the bamboo tube). For each trial, we feed the real video sequence into GC-IDM, infer the corresponding gripper-centric action sequence, and replay it in the real world to measure Replay Execution Accuracy. As shown in Table~\ref{tab:relpay_results}, we showcased four of the tasks, GC-IDM substantially outperforms standard baselines such as ResNet-based inverse dynamics model and AVDC~\cite{Ko2023Learning} across all tasks, achieving an overall replay accuracy of 90\%, demonstrating strong fidelity in mapping real videos to executable actions. Representative frames from ground-truth video replays are provided in Figure~\ref{fig:real-world}.

This high replay accuracy confirms that GC-IDM is a reliable evaluator of physical plausibility. Consequently, when GC-IDM fails on model-generated videos, we can attribute the failure to deficiencies in video realism rather than inaccuracies in the IDM itself.

\begin{table}[h!]
\centering
\caption{\textbf{Comparison of Success Rate (\%) on various tasks across different models on ground-truth video replay.}}
\label{tab:relpay_results}
\resizebox{\linewidth}{!}{%
\begin{tabular}{@{\hskip 5pt}l@{\hskip 5pt} @{\hskip 5pt}cccc@{\hskip 5pt}}
\toprule
\textbf{Model / Task} & \textbf{Bread to Plate} & \textbf{Close Drawer} & \textbf{Move Milk} & \textbf{Bread in Drawer} \\
\midrule
\textbf{ResNet-MLPs}     & 6/10 & 7/10 & 5/10 & 4/10  \\
\textbf{AVDC}~\cite{Ko2023Learning}        & 3/10 & 4/10 & 4/10 & 2/10 \\
\rowcolor{mylightgreen}
\textbf{GC-IDM}~\cite{chi2025wow}   & \textbf{10/10} & \textbf{9/10} & \textbf{9/10} & \textbf{8/10} \\
\bottomrule
\end{tabular}
}
\end{table}

\section{Case Results Visualization}
\label{sec:more_visualizations}
\begin{figure*}[!h]
  \centering
  \includegraphics[width=1.0\textwidth]{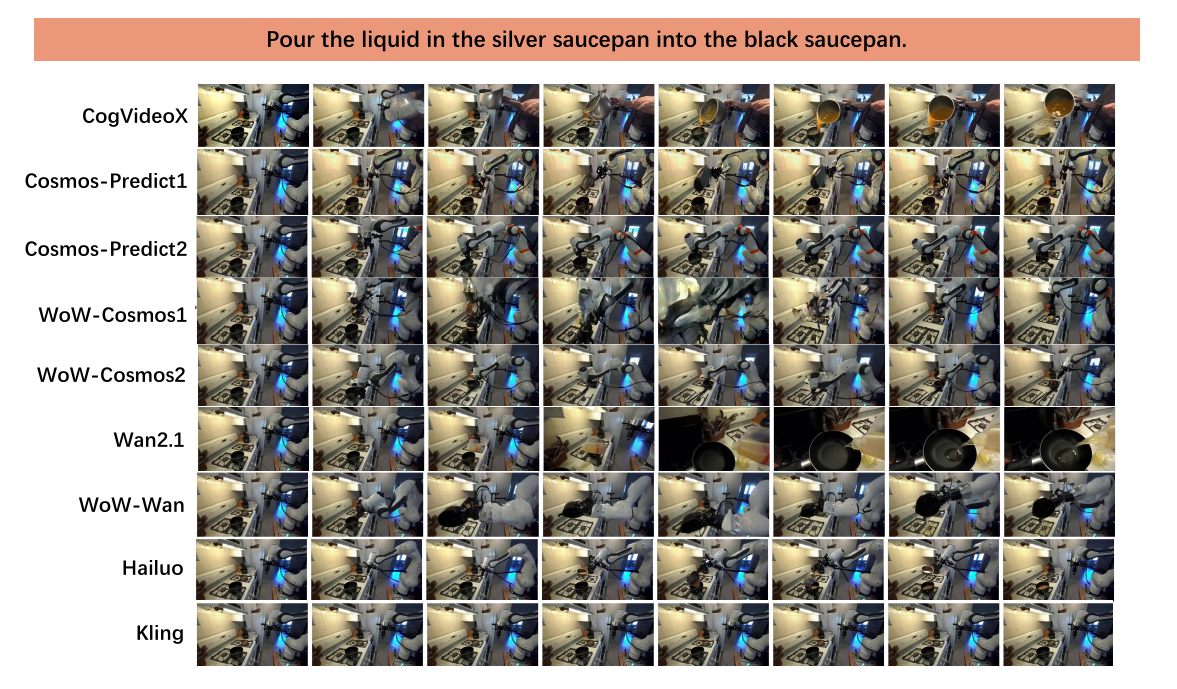} 
  \caption{\textbf{More Case Visualization in Perception.} }
  \label{fig:visualization2}
\end{figure*}

\begin{figure*}[!h]
  \centering
  \includegraphics[width=1.0\textwidth]{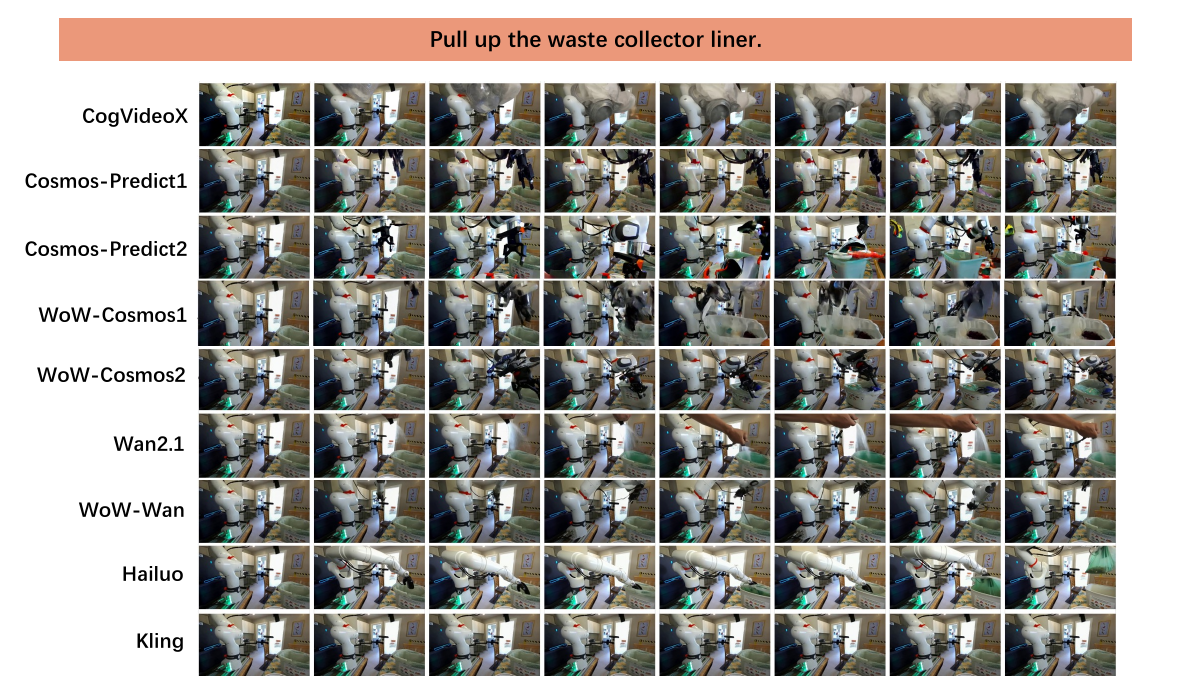} 
  \caption{\textbf{More Case Visualization in Prediction.} }
  \label{fig:visualization3}
\end{figure*}

\begin{figure*}[!h]
  \centering
  \includegraphics[width=1.0\textwidth]{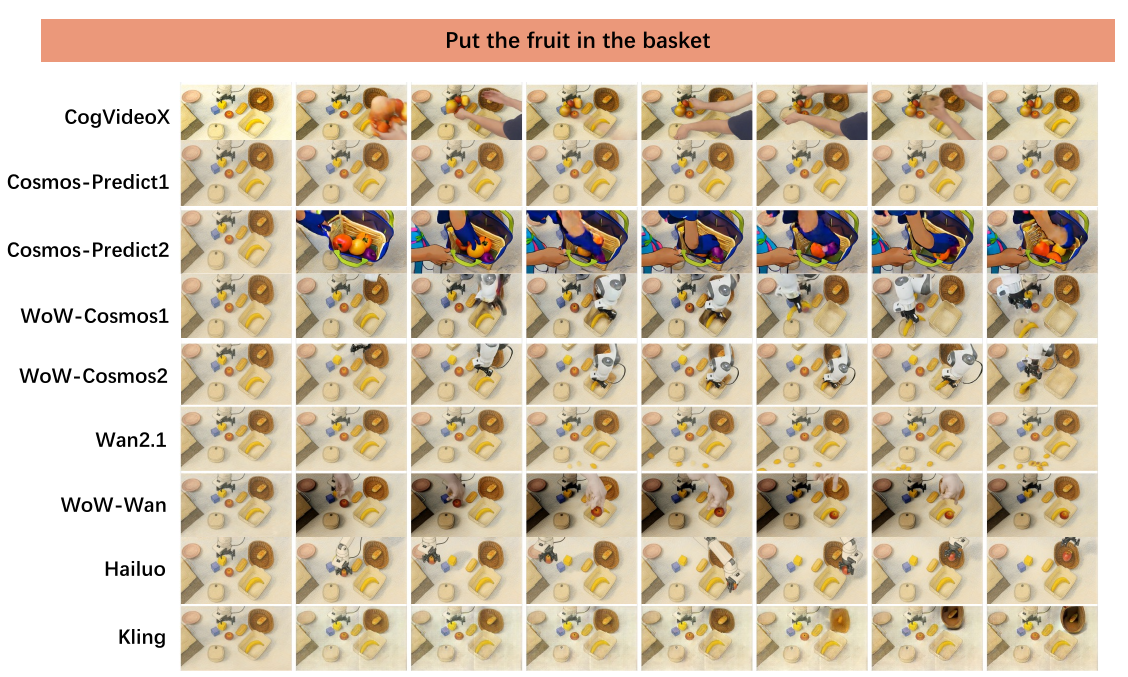} 
  \caption{\textbf{More Case Visualization in Generalization.} }
  \label{fig:visualization4}
\end{figure*}

We also collect more generated results of different models for comparision, as visualized in Figure~\ref{fig:visualization2}, ~\ref{fig:visualization3}, ~\ref{fig:visualization4}.

\section{Prompts}
For the prompts that we used in GPT for evaluation and data selection, or our own trained Physical MLLM.

\tcbset{
  promptstyle/.style={
    colback=yellow!15,    
    colframe=yellow!15,   
    boxrule=0pt,          
    left=10pt,
    right=10pt,
    top=10pt,
    bottom=10pt,
    boxsep=5pt,
  }
}


\centering

\begin{tcolorbox}[
    promptstyle, 
    width=\linewidth, 
    breakable
] 
    \subsection*{GPT Data Selection Prompt}

    \begin{lstlisting}[
        language=Tex, 
        basicstyle=\small\ttfamily,
        breaklines=true,
        frame=none, 
        columns=fullflexible,
        backgroundcolor=\color{yellow!15}, 
    ]
You are a robotics data selection and instruction rewriting assistant.
Input: one task instruction (text) and one first-frame image.
Output: JSON only, describing which data dimensions this sample belongs to, with evidence, confidence scores, and a rewritten instruction only for Perception dimensions.

Absolute Rules

Use only what is visible in the first frame and what is stated in the instruction. Do not hallucinate or import outside knowledge.

A dimension is valid only if:

The image provides clear visual evidence, and

The original instruction can be unambiguously rewritten (if Perception) or directly preserved (if Prediction/Planning).

If evidence is insufficient or ambiguous, do not assign the dimension; instead, explain in issues.

Multiple dimensions can be assigned; each must include its own score and evidence.

Instruction rewriting requirement:

Perception including rewrite instruction into an equivalent form using attributes / functions / spatial relations / affordances.

Prediction and Planning will always keep the original instruction unchanged.

Dimension Checklist
A. Perception

object-centric / object-attribute (color, number, shape, size, type)
    Paths must follow the format:
    'Perception/object-centric/object-attribute/color'
    'Perception/object-centric/object-attribute/type'
    etc.

Rewrite into attribute-based description.

object-centric / object-function

Rewrite into function-based description.

scene-centric / spatial-relation

Rewrite into relation-based description.

affordance-centric / object-affordance

Rewrite into affordance-based description.

B. Prediction

camera-view including no-occlude | semi-occlude.

physical-interaction including single-object/rigid | single-object/fluid | single-object/deformable | single-object/articulated | multi-object/rigid-rigid | multi-object/rigid-fluid | multi-object/rigid-deformable | multi-object/fluid-fluid | dual-arm-cooperate 

corner-case.

Keep original instruction (no rewriting).

C. Planning

long-term-instruction.

Keep original instruction (no rewriting).

JSON Output Format
{
  "input_instruction": "<original instruction>",
  "dimensions": [
    {
      "path": "Perception/object-centric/object-attribute/color",
      "score": 0.9,
      "evidence": "blue plate, red tomato, yellow pepper visible",
      "rewrite": "move the red and yellow objects to the blue object"
    },
    {
      "path": "Prediction/camera-view",
      "score": 0.8,
      "evidence": "all target objects fully visible",
      "rewrite": "keep original instruction"
    }
  ],
  "prediction": {
    "camera-view": {
      "label": "no-occlude | semi-occlude | unknown",
      "evidence": "<short evidence>",
      "score": 0.0
    },
    "physical-interaction": [
      {
        "label": "single-object/rigid | ...",
        "evidence": "<short evidence>",
        "score": 0.0
      }
    ]
  },
  "planning": [
    {
      "label": "long-term-instruction",
      "evidence": "<short evidence>",
      "score": 0.0,
      "rewrite": "keep original instruction"
    }
  ],
  "issues": [
    "<if there is missing evidence or ambiguity, explain here>"
  ]
}


For Perception dimensions, rewrite contains the modified instruction.

For Prediction and Planning dimensions, rewrite must be exactly "keep original instruction".

score is confidence for 0 to 1.

evidence is short, factual, visual clues only.

**Example**

Input:

instruction: move tomato and pepper to the plate

image: blue plate, red tomato, yellow pepper; no occlusion.

Output:

{
  "input_instruction": "move tomato and pepper to the plate",
  "dimensions": [
    {
      "path": "Perception/object-centric/object-attribute/color",
      "score": 0.95,
      "evidence": "blue plate; red tomato; yellow pepper visible",
      "rewrite": "move the red and yellow objects to the blue object"
    },
    {
      "path": "Prediction/camera-view",
      "score": 0.9,
      "evidence": "all target objects fully visible",
      "rewrite": "keep original instruction"
    }
  ],
  "prediction": {
    "camera-view": {
      "label": "no-occlude",
      "evidence": "all objects clearly visible",
      "score": 0.9
    },
    "physical-interaction": [
      {
        "label": "single-object/rigid",
        "evidence": "rigid tabletop objects",
        "score": 0.8
      }
    ]
  },
  "planning": [],
  "issues": []
}
    \end{lstlisting}
    
\end{tcolorbox}

\centering

\begin{tcolorbox}[
    promptstyle, 
    width=\linewidth, 
    breakable
] 
    \subsection*{Instruction Semantic Correctness: Extract Video Semantic Caption}

    \begin{lstlisting}[
        language=Tex, 
        basicstyle=\small\ttfamily,
        breaklines=true,
        frame=none, 
        columns=fullflexible,
        backgroundcolor=\color{yellow!15}, 
    ]
Please analyze the instruction that I used to guide the model to generate video and the content of the following groundtruth video(there maybe no groundtruth video, if so please use only the instruction), extract semantic descriptions related to the robot's behavior. Focus on the following five aspects and provide one clear, concise sentence for each:
1. Initial State: The state of the object(s) and environment before any action (e.g., position, posture, quantity)
2. Processing State: The interaction process while the robot arm is operating (e.g., how it moves or manipulates the object)
3. Final State: The outcome after the action (e.g., new position or state of the object, whether the task goal was achieved)
4. Action: The main type of operation performed (e.g., grasping, pushing, placing)
5. Object: The object being manipulated (e.g., its color, shape, category)

Please output your result in the following format:
```
Initial State: ...
Processing State: ...
Final State: ...
Action: ...
Object: ...
```

Guidelines:
- Keep each sentence objective, specific, and based only on what is visible in the video.
- Do not add assumptions or imagined information.
- If any aspect is unclear or not visible, write "Unknown" for that field.
- Avoid overly abstract descriptions such as "successfully completed the task."

    \end{lstlisting}
    
\end{tcolorbox}

\centering

\begin{tcolorbox}[
    promptstyle, 
    width=\linewidth, 
    breakable
] 
    \subsection*{Instruction Semantic Correctness: Compare Video Semantic Caption}

    \begin{lstlisting}[
        language=Tex, 
        basicstyle=\small\ttfamily,
        breaklines=true,
        frame=none, 
        columns=fullflexible,
        backgroundcolor=\color{yellow!15}, 
    ]
You are given structured semantic captions from two videos: the original ground truth and a generated version. Compare them across the following five dimensions and evaluate their semantic consistency.
"comparison_dimensions": {    
    "initial_state": "The object/environment status before any action",
    "processing_state": "How the robot interacts with the object during operation", 
    "final_state": "The outcome or resulting object state",
    "action": "The type of manipulation performed",
    "object": "The object being manipulated"
},
"scoring_guidelines": {
    "level_1": {
        "score": 1,
        "description": "Fully Consistent, Descriptions convey the same semantic content with only stylistic or wording differences"    
    },
    "level_2": {
        "score": 0.5,
        "description": "Partially Consistent, Descriptions differ in minor semantic details or omit secondary information, but task intent remains aligned"
    },
    "level_3": {
        "score": 0,
        "description": "Inconsistent, Descriptions conflict semantically, refer to different actions, goals, or object properties"
    }
},
"output_format": {
    "initial_state": "Score = [0 / 0.5 / 1] - Reason: ...",
    "processing_state": "Score = [0 / 0.5 / 1] - Reason: ...", 
    "final_state": "Score = [0 / 0.5 / 1] - Reason: ...",
    "action": "Score = [0 / 0.5 / 1] - Reason: ...",
    "object": "Score = [0 / 0.5 / 1] - Reason: ...",
    "overall": "Score = [0 / 0.5 / 1] - Summary Reason: ..."
},
"evaluation_principles": [
"Focus strictly on the semantic content of the captions, do not infer or assume actions or states that are not explicitly stated.",
"The overall score may reflect the average of the five dimensions, but should also consider whether the task intent and outcome remain semantically aligned."
]
Ground truth video caption is: {ground_truth_caption}
Generated video caption is: {generated_caption}
    \end{lstlisting}
    
\end{tcolorbox}

\centering

\begin{tcolorbox}[
    promptstyle, 
    width=\linewidth, 
    breakable
] 
    \subsection*{Instruction Semantic Correctness: Evaluate Video Action Execution}

    \begin{lstlisting}[
        language=Tex, 
        basicstyle=\small\ttfamily,
        breaklines=true,
        frame=none, 
        columns=fullflexible,
        backgroundcolor=\color{yellow!15}, 
    ]
You are given a language instruction describing a series of robot manipulation actions, and a generated video that attempts to follow that instruction.
Your task has two parts:
PART 1: Action Sequence Consistency (Proportional Scoring)
Parse the instruction and extract the intended object-action pairs in order.
Analyze the generated video to determine the actual order of the executed object-action pairs. Compare the two sequences and compute a sequence match score:
    Sequence Match Scoring: 
        Score = (number of matching object-action pairs in correct order) / (total number of instruction pairs)
        Matching means: the object-action pair appears in the video and in the same relative order as in the instruction. Partial matches or out-of-order actions reduce the score.
    Output format:
        Instruction Sequence: [(object1, action1), (object2, action2), ...]
        Video Sequence: [(object1, action1), (object2, action2), ...]
        Sequence Match Score: X.XX (range: 0.00-1.00) - Reason: ...
        
PART 2: Action Execution Quality (Numerical Evaluation)
For each (object, action) pair in the instruction, evaluate execution quality using a 1-5 numerical scale:
    Execution Quality Scoring:
        1 - Object did not move AND action was not attempted
        2 - Object moved slightly OR action was partially attempted
        3 - Object moved slightly AND action was partially attempted
        4 - Object reached target state OR action completed successfully
        5 - Object reached target state AND action completed successfully

    Base this strictly on visible evidence in the video. Be conservative in scoring.

    Output format:
        Execution Quality:
        - (object1, action1): Score = [1-5] - Reason: ...
        - (object2, action2): Score = [1-5] - Reason: ...
        
Final Combined Output:
    Instruction Sequence: ...
    Video Sequence: ...
    Sequence Match Score: X.XX (range: 0.00-1.00) - Reason: ...
    Execution Quality:
    - (object1, action1): Score = [1-5] - Reason: ...
    - (object2, action2): Score = [1-5] - Reason: ...
    Instruction is: {instruction}

    \end{lstlisting}
    
\end{tcolorbox}

\centering
\begin{tcolorbox}[
    promptstyle, 
    width=\linewidth, 
    breakable 
] 
    \subsection*{Physical Common Sense: Stage 2: Scoring Alignment Fine-tuning Prompt }

    \begin{lstlisting}[
        language=Tex, 
        basicstyle=\small\ttfamily,
        breaklines=true,
        frame=none, 
        columns=fullflexible,
        backgroundcolor=\color{yellow!15}, 
    ]
You are an expert AI video evaluator. Your task is to analyze the provided video, which was generated from the accompanying text prompt. You must score the video on a scale of 1 to 5 across four dimensions based on the precise criteria listed below.

Evaluation Criteria:

1. Video Quality (1-5):

    5: Clear, stable, natural colors, no generation artifacts.

    4: Minor flaws (e.g., occasional flicker, slight blur) that don't detract from the overall view.

    3: Noticeable flaws (e.g., persistent noise, subject out of focus) but core content is understandable.

    2: Severe defects (e.g., unstable subject form, heavy jitter) that significantly impair understanding.

    1: Extremely poor quality, making the subject or content unrecognizable.

2. Instruction Following (1-5):

    5: Perfectly matches the prompt; all core elements and details are accurately represented.

    4: Core elements are correct, but a minor detail (e.g., color, location) is missed or incorrect.

    3: The core element (subject or action) deviates significantly but is still partially related to the prompt.

    2: The core element is fundamentally misinterpreted; the content is seriously inconsistent with the prompt.

    1: Completely unrelated to the prompt; a random generation.

3. Physical Consistency (1-5):

    5: Fully consistent with physics; interactions are believable, no object clipping or floating.

    4: A momentary, almost unnoticeable physical anomaly (e.g., slight clipping) that doesn't affect the task.

    3: A noticeable physical error (e.g., arm passes through a table) that doesn't break the core task logic.

    2: A severe physical error that causes the task to fail or its logic to break (e.g., a hand passes through the target object).

    1: A complete lack of physical logic; chaotic and disorderly interactions.

4. Planning Logic (1-5):

    5: The plan is logical, efficient, with no redundant actions, and successfully achieves the goal.

    4: The plan is effective and successful but contains minor redundant or inefficient movements.

    3: The goal is achieved, but the process is convoluted or clumsy; the plan is illogical.

    2: The plan has critical errors or missing steps, leading to task failure.

    1: Actions are random and chaotic, with no discernible goal or plan.

Analyze the following inputs:

Original Prompt: {instruction}
Generated Video: <video>
Your final answer must be a single, valid JSON object and nothing else. Do not include any explanations, conversational text, or markdown formatting around the JSON block.
Your JSON Answer:
{
  "video_quality": <integer_score>,
  "instruction_following": <integer_score>,
  "physical_consistency": <integer_score>,
  "planning_logic": <integer_score>
}
    \end{lstlisting}
    
\end{tcolorbox}


\centering

\begin{tcolorbox}[
    promptstyle, 
    width=\linewidth, 
    breakable
] 
    \subsection*{Physical Common Sense: 6-Dimension Evaluation Prompt}

    \begin{lstlisting}[
        language=Tex, 
        basicstyle=\small\ttfamily,
        breaklines=true,
        frame=none, 
        columns=fullflexible,
        backgroundcolor=\color{yellow!15}, 
    ]
Please evaluate the following video strictly based on whether it follows real-world physics laws.
This is not a test of whether it is AI-generated, but only judge its physical plausibility.
Scoring System (1 to 5 scale):
For each of the six categories below, assign a score from 1 to 5:
5 = Fully consistent with real-world physics
4 = Mostly consistent, only small flaws
3 = Partially consistent, with some clear issues
2 = Often inconsistent, unrealistic in multiple ways
1 = Clearly violates physical laws
- null = Not applicable (no relevant content in video)
For each of the following categories:
- Give a score (1 to 5) or null if the category does not appear at all in the video
- Provide a brief explanation (1 to 2 sentences) for each score
Return your output in structured JSON as shown below.

---
Evaluation Criteria (Detailed):
1. Object Interaction and State Changes
  - Do objects respond naturally when they touch, collide, bounce, break, or are pushed/pulled?
  - Is there a realistic cause-and-effect in object contact?
  - Do interactions follow Newton's laws (e.g., action/reaction, inertia)?
  - Are deformations (e.g., squashing, bending) consistent with materials?
  - Do objects maintain consistent color, quantity, shape, and size throughout interactions?
2. Basic Physical Properties
  - Does gravity act in the correct direction and magnitude (e.g., falling speed)?
  - Is there realistic motion under inertia and deceleration?
  - Does friction behave naturally (e.g., sliding slows over time)?
  - Are materials (heavy/light) behaving correctly based on mass or density?
3. Temporal and Causal Consistency
  - Do effects follow causes in the right order and with realistic timing?
  - Are there appropriate delays in mechanical systems, collisions, or triggered motions?
  - Is there a consistent flow of time across all elements (e.g., no time jumps or impossible overlaps)?
4. Lighting, Shadows, and Reflections
  - Are shadows cast in the correct direction and shape relative to light sources?
  - Is there consistent lighting across surfaces and objects?
  - Do reflective surfaces (mirrors, metal, water) behave correctly based on angles?
  - Is light intensity and falloff physically accurate (not flat or overly uniform)?
5. Fluid and Particle Behavior
  - Do water, smoke, fire, or dust behave according to fluid dynamics?
  - Is there natural turbulence, diffusion, and randomness?
  - Do particles (like debris, sparks, splashes) move in a physically plausible way?
  - Is there appropriate interaction with solid surfaces (e.g., splashes bounce, smoke drifts)?
6. Local Anomalies or Violations
  - Are there any visual inconsistencies like:
    - sudden object movement without cause
    - teleportation or disappearing items
    - objects passing through others (no collision)
    - animation glitches, floating artifacts, or broken motion?
  - Do all parts of the scene stay consistent frame to frame?
---
Return your output in structured JSON as shown below.
    - object_interaction: {score: int|null, comment: str}
    - physical_properties: {score: int|null, comment: str}
    - temporal_consistency: {score: int|null, comment: str}
    - lighting_and_reflections: {score: int|null, comment: str}
    - fluids_and_particles: {score: int|null, comment: str}
    - local_anomalies: {score: int|null, comment: str}
    \end{lstlisting}
    
\end{tcolorbox}


\centering

\begin{tcolorbox}[
    promptstyle, 
    width=\linewidth, 
    breakable
] 
    \subsection*{Dense Prompts Extension Prompt}

    \begin{lstlisting}[
        language=Tex, 
        basicstyle=\small\ttfamily,
        breaklines=true,
        frame=none, 
        columns=fullflexible,
        backgroundcolor=\color{yellow!15}, 
    ]
You are an expert in Embodied AI tasked with generating dense and standardized textual descriptions of robot episodes. Your goal is to re-caption a brief, potentially ambiguous, episode description  into a richer, more precise narrative,but no more than 150 words. This serves two main purposes:
1.  To clarify ambiguous action descriptions with precise technical terminology.
2.  To unify the textual representation of similar actions across different datasets, ensuring consistency in phrasing and detail.

When generating the dense caption, adhere to the following structure and include these specific details:

1.  **Overall Scene & Goal Overview:**
    *   Begin with a concise summary of the robot's primary goal or the main action being performed in the scene, directly derived or inferred from the input .

2.  **Environment Description:**
    *   Describe the robot's operational environment (e.g., kitchen, laboratory, office, cluttered tabletop, outdoor path).
    *   Include relevant ambient conditions: lighting (e.g., bright, dim, natural, artificial), indoor/outdoor setting, and weather if applicable (e.g., sunny, overcast for outdoor scenes).

3.  **Robot Characteristics:**
    *   Specify key characteristics of the robot if inferable or if it's a common type in embodied AI (e.g., humanoid, mobile manipulator with a single arm, quadruped, drone).
    *   Note its general appearance if significant (e.g., color scheme, predominant material like metallic or plastic).
    *   Mention its gripper type if relevant to the action (e.g., parallel jaw gripper, suction cup, multi-finger dexterous hand).

4.  **Camera Perspective & Motion:**
    *   Detail the camera's perspective (e.g., first-person (egocentric/robot's POV), third-person fixed, third-person dynamic/following, eye-in-hand, overhead).
    *   Describe any significant camera movement (e.g., static, panning to follow action, smooth tracking shot, handheld jitter if applicable).

5.  **Robot Sub-goal & Granular Action Breakdown:**
    *   Infer and explicitly state the immediate sub-goal the robot is trying to achieve based on the input.
    *   Provide a granular breakdown of the robot's actions. Use precise, standardized technical action terminology (e.g., "Navigate_to [location]", "Perceive [object]", "Grasp [object]", "Lift [object]", "Move_arm_to [target_pose]", "Place [object] at [location]", "Release [object]", "Push [object]", "Rotate [object]").
    *   Clearly identify the object(s) being manipulated and any changes in their state or location.

6.  **Post-Action State:**
    *   Briefly describe the state of the robot and the key manipulated object(s) *after* the described action sequence is completed.

**Constraints:**
*   The final description should be dense, focusing on factual and observable details.
*   Aim for a total length of approximately 100-150 words.
*   Use consistent terminology for common actions and objects.
*   Avoid filler words or subjective stylistic descriptions not grounded in typical embodied AI contexts.
*   Only describe elements plausibly present or inferable from a typical embodied AI episode matching the `Given Input`. Do not invent fantastical details.
*   Never mention elements outside the implied scene or action.

---
**Example of Use:**

Let's say the given short episode description  is: "drop the apple in the bowl"

**Given Input :drop the apple in the bowl

**Your Dense Recaption (Output based on the prompt above):**

The robot's primary goal is to place an apple into a designated bowl.
The scene is set in a brightly lit kitchen environment, specifically on a clean countertop. The robot is a single-arm manipulator, predominantly white with grey joints, equipped with a parallel jaw gripper, likely mounted on a stationary or mobile base (partially visible). The camera is a static third-person view, positioned to offer a clear side-angle of the robot's arm and the workspace, including the target bowl.
The robot's sub-goal is to successfully deposit the apple. The action sequence begins with the robot's arm, already grasping a red apple, positioned directly above a ceramic bowl. The arm executes a 'Move_arm_to [bowl_center_above]' command, followed by a 'Lower_arm' motion. The gripper then performs a 'Release [apple]' action. Finally, the arm might execute a 'Retract_arm' motion away from the bowl.
After the action, the red apple rests inside the bowl on the countertop, and the robot's gripper is open and clear of the bowl.

Given Input : {instruction}
Your Dense Recaption:

    \end{lstlisting}
    
\end{tcolorbox}

\clearpage


\end{document}